\documentclass[10pt,twocolumn,letterpaper]{article}

\usepackage[pagenumbers]{cvpr} 
\usepackage{bm}
\usepackage{comment}

\usepackage[dvipsnames]{xcolor}

\usepackage{stmaryrd}
\usepackage{textcomp}

\newcommand{\iverson}[1]{\left \llbracket #1 \right \rrbracket}

\definecolor{cvprblue}{rgb}{0.21,0.49,0.74}
\usepackage[pagebackref,breaklinks,colorlinks,citecolor=cvprblue]{hyperref}

\usepackage{graphicx} 
\usepackage{verbatim}
\usepackage{multirow}
\usepackage[accsupp]{axessibility} 

\def\paperID{5051} 
\def\confName{CVPR}
\def\confYear{2024}

\title{Multiway Point Cloud Mosaicking with Diffusion and Global Optimization}

\author{Shengze Jin$^1$ \quad Iro Armeni$^2$ \quad Marc Pollefeys$^{1,3}$ \quad D\'{a}niel Bar\'{a}th$^1$\\
$^1$ Department of Computer Science, ETH Zurich, Switzerland\\ 
$^2$ Department of Civil and Environmental Engineering, Stanford University\\  
$^3$ Microsoft Mixed Reality \& AI Lab, Zurich, Switzerland\\
}

\begin{document}
\maketitle
\begin{abstract}
We introduce a novel framework for multiway point cloud mosaicking (named Wednesday), designed to co-align sets of partially overlapping point clouds -- typically obtained from 3D scanners or moving RGB-D cameras -- into a unified coordinate system.
At the core of our approach is ODIN, a learned pairwise registration algorithm that iteratively identifies overlaps and refines attention scores, employing a diffusion-based process for denoising pairwise correlation matrices to enhance matching accuracy. 
Further steps include constructing a pose graph from all point clouds, 
performing rotation averaging, a novel robust algorithm for re-estimating translations optimally in terms of consensus maximization and translation optimization.
Finally, the point cloud rotations and positions are optimized jointly by a diffusion-based approach. 
Tested on four diverse, large-scale datasets, our method achieves state-of-the-art pairwise and multiway registration results by a large margin on all benchmarks. 
Our code and models are available at
\href{https://github.com/jinsz/Multiway-Point-Cloud-Mosaicking-with-Diffusion-and-Global-Optimization} {https://github.com/jinsz/Multiway-Point-Cloud-Mosaicking-with-Diffusion-and-Global-Optimization}.
\end{abstract}    
\vspace{-7mm}
\section{Introduction}
\label{sec:intro}

Registering multiple partially overlapping 3D point cloud fragments into a unified coordinate system is crucial to comprehensively representing an environment.  
This procedure has a wide range of applications in computer vision and robotics, such as in 3D scene understanding~\cite{jatavallabhula2023conceptfusion,Takmaz2023openmask3d}, augmented reality~\cite{taguchi2013point,sarlin2022lamar}, and autonomous driving~\cite{wolcott2014visual,lu2019l3,palieri2020locus}.
In particular, LiDAR or RGB-D-based mapping is often employed to build large-scale maps in self-driving and mobile robotics due to their direct and accurate 3D point cloud sensing capability. There are typically two steps in building such maps: pairwise and multiway registration.

\begin{figure}[t]
    \begin{center}
    \includegraphics[width=0.9\linewidth,trim={3.3cm 12.5cm 6.5cm 3cm},clip]{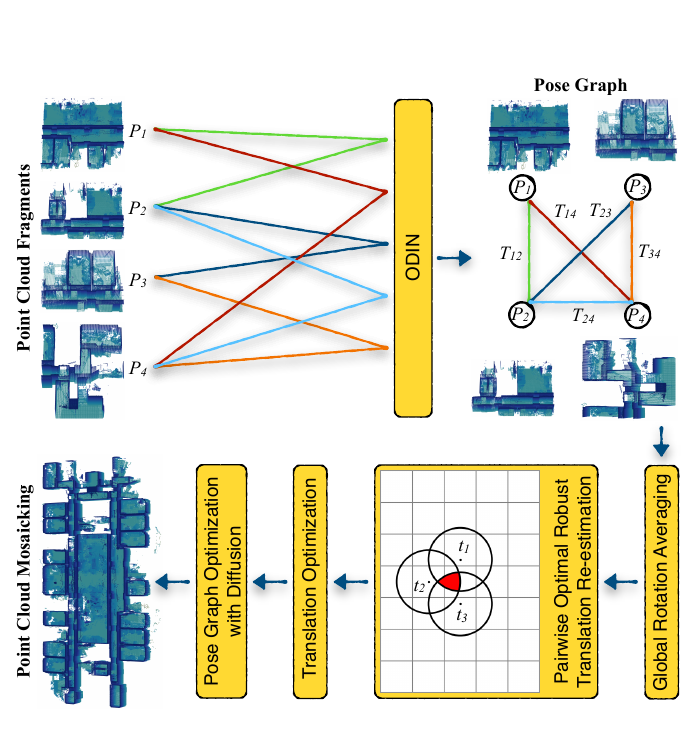}
    \end{center}
    \vspace{-5mm}
    \caption[Teaser Image]{
    The proposed multiway registration method, Wednesday, starts with pairwise registration of an unordered set of partially overlapping point clouds using the proposed matcher (ODIN). 
    The process then optimizes the constructed pose graph, which includes global point cloud poses (vertices) and relative transforms (edges), through a sequence of steps: (a) global rotation averaging, (b) a novel optimal robust translation re-estimation method conceptualized as finding maximal sphere overlaps, (c) position averaging, and (d) diffusion-based pose graph optimization.
    The output is the point clouds in a unified coordinate system. 
    }
    \label{fig:teaser}
    \vspace{-5mm}
\end{figure}

The pairwise registration of partially overlapping point clouds is a thoroughly investigated problem, with several methods proposed over time. 
Conventional approaches to the pairwise problem are based on imposing geometric constraints~\cite{rabbani2007integrated,zeisl2013automatic,theiler2015globally} on hand-engineered feature descriptors~\cite{johnson1999using,flint2007thrift,rusu2009fast,tombari2010unique} employing robust estimators~\cite{fischler1981random,bustos2017guaranteed,barath2018graph}. 
In recent years, research on local descriptors for pairwise registration of 3D point clouds has shifted towards deep learning methodologies~\cite{khoury2017learning,zeng20173dmatch,yew20183dfeat,deng2018ppfnet,deng2018ppf,lu2019deepvcp,gojcic2019perfect,huang2021predator,qin2023geotransformer,yu2023peal,yu2023rotation}.
Such approaches have demonstrated impressive results by implicitly learning local and global scene characteristics, which are then distilled into highly distinctive local descriptors.
Although these methods have proven to be effective, directly applying them to the multiway problem has conceptual drawbacks: (i) low overlap between adjacent point clouds may result in incorrect matches, and (ii) the reliance only on local evidence, which can be problematic in scenes with scarce or repetitive 3D structures.

In contrast to pairwise registration, point cloud mosaicking (\ie, the globally consistent multiway alignment of unorganized point clouds) has received much less attention. Traditional approaches~\cite{endres20133,whelan2013deformation,xiao2013sun3d,zhou2013dense,steinbrucker2013large,choi2015robust,chen2023deepmapping2} primarily tackle this problem from the robotics perspective in scenarios where the differences between adjacent point clouds are minimal (\eg, they come from subsequent RGB-D frames), and the recorded data distinctly reveals the trajectory of the robot as it captures the scene. 
In such cases, pairwise registration provides accurate initialization for the multiway problem, with the trajectory providing additional constraints~\cite{zhou2016fast} that can be incorporated into the optimization process.
Other methods~\cite{govindu2004lie,torsello2011multiview,arie2012global,arrigoni2014robust,bernard2015solution,arrigoni2016spectral,maset2017practical,birdal2018bayesian}, focusing on having unordered point cloud sets, frame the challenge as an optimization procedure. However, in practice, it is still sensitive to the failures in the pairwise registration. 

Recent progress in the field has seen the introduction of end-to-end  pipelines~\cite{gojcic2020learning,yew2021learning,wang2023robust,chen2023deepmapping2} that aim at learning specific local and global characteristics of the scene.
However, such methods prioritize ease of training over efficacy during inference. 
To facilitate differentiability, the representations (\eg, rotation manifold) and algorithms (\eg, iteratively reweighted least-squares) utilized are selected for their compatibility with the end-to-end pipeline rather than for their potential to yield optimal performance at inference. 

This paper focuses on designing a pipeline for accurate point cloud registration even in challenging environments with spatial and temporal changes.
We enhance state-of-the-art pairwise registration algorithms based on two observations: (i) the predicted matching matrix often contains noise, and its denoising leads to improved 3D-3D matches; (ii) while finding individual 3D point matches is key to estimating the rigid transformation,
the underlying objective is to find the best point cloud overlap. 
This can be directly measured and integrated into the matching process.  
To achieve accurate multiway registration, we rely on classical geometric optimization-based approaches known for their accuracy and generalizability.
The proposed pipeline is the result of carefully selected and \textit{new} optimization techniques for the best accuracy, combining learning-based and classical algorithms to benefit from data-driven approaches while maintaining the efficiency and applicability of geometric methods.
The contributions are as follows:
\begin{itemize}
    \item A novel pipeline (see Fig.~\ref{fig:teaser}) for multiway point cloud registration consisting of modules for pairwise estimation, global rotation averaging, translation re-estimation and averaging, and diffusion-based final optimization.
    \item A novel pairwise point cloud registration method, ODIN, incorporating point cloud overlap scores into attention learning and diffusion-based correlation matrix denoising for highly accurate pairwise matching. 
    \item An efficient and globally optimal robust consensus (\ie, inlier number) maximization approach for re-estimating relative translations given known global orientations.
    \item As a technical contribution, we adapt a recent diffusion-based pose graph optimization~\cite{wang2023posediffusion} to point clouds. 
\end{itemize}
The proposed advancements and other methods fused into a single pipeline achieve state-of-the-art accuracy by a \textit{great} margin. It achieves 82\% rotation error reduction on the most challenging dataset~\cite{nss2023}. It also reduces the average position error by 27\% across the tested datasets. 

\section{Related Work}
\label{sec:Re_work}

\begin{figure*}[t]
\begin{center}
    \includegraphics[angle=-90,width=0.95\textwidth,trim={7.5cm 1cm 5.5cm 3cm},clip]{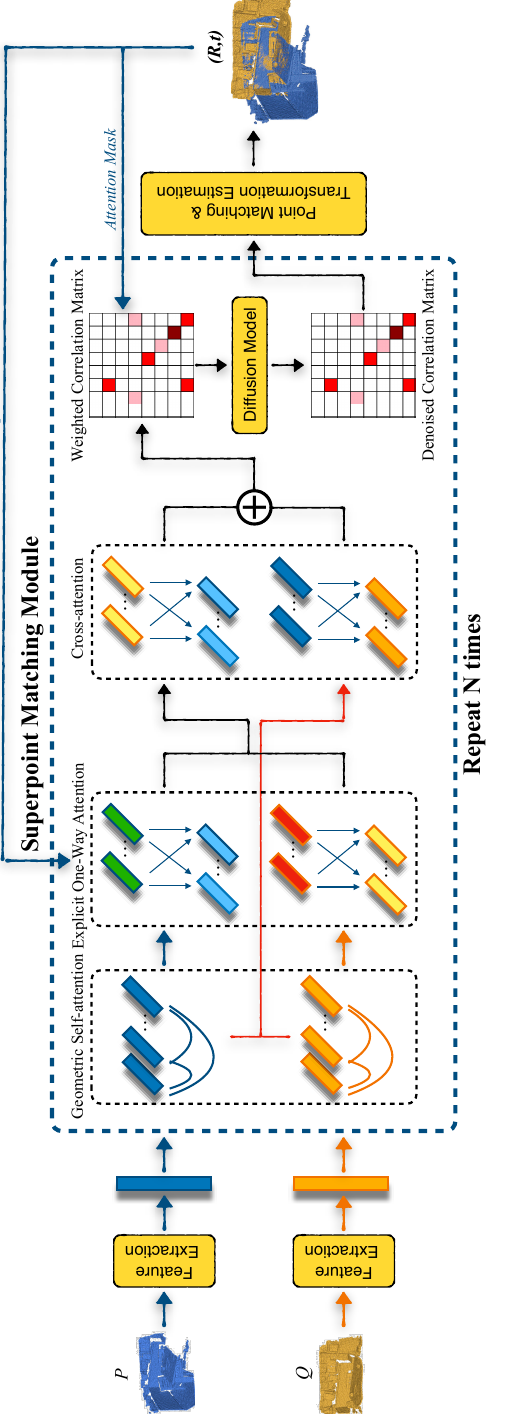}
\end{center}
    \vspace{-0.6cm}
    \caption[ODIN pipeline]{\textbf{Two-view registration}. Given two points clouds as input, ODIN (Section~\ref{sec:pairwise}) first extracts features that are then processed by geometric self-attention to learn point-specific attention features. 
    Next, the process is separated into two parallel streams: 
    In (a), the features are processed by explicit one-way self and cross-attentions. This process incorporates overlap scores determined in the final stage. 
    In (b), the features directly go through cross-attention. The determined correlation matrix is the weighted average of the correlations from the two streams. A diffusion-based denoising cleans the correlations. Finally, point matching and transformation estimation are performed. The overlap scores implied by the estimated transform are sent back to the attention learning module as a mask and the process starts over.  }
    \vspace{-4mm}
\end{figure*}

\noindent
\textbf{Pairwise registration} is traditionally a two-step process. 
The first step is the coarse alignment stage, where initial estimates of the relative transformations are obtained. 
The second step is the refinement stage, where the global poses are iteratively refined to minimize the 3D registration error, assuming a rigid transformation. 
Coarse alignment often uses handcrafted~\cite{johnson1999using,flint2007thrift,rusu2009fast,tombari2010unique} or learned~\cite{huang2021predator,qin2023geotransformer,yu2023peal,yu2023rotation} 3D local feature descriptors to establish tentative pointwise correspondences.
They are used with a RANSAC-like robust estimator~\cite{fischler1981random} or geometric hashing~\cite{drost2010model,birdal2015point,hinterstoisser2016going} to find the pose parameters and the consistent matches. 
Another approach uses 4-point congruent sets to establish correspondences~\cite{aiger20084,mellado2014super,theiler2014keypoint}. 
In the refinement stage, coarse transformation parameters are fine-tuned using a variant of the iterative closest point (ICP) algorithm~\cite{besl1992method}. 
However, ICP-like algorithms~\cite{chetverikov2002trimmed,li20073d,yang2015go} are not robust against outliers. 
ICP algorithms can be extended to use additional radiometric, temporal, or odometry constraints~\cite{zhou2016fast}. 

Recent work~\cite{wang2019deep,lu2019deepvcp,huang2021predator,yu2023peal,jin2023q} either directly regress transformations or refine correspondences by considering the information from both point clouds, \eg, with attention layers.
Our proposed pairwise method falls into this category, building upon transformer-based alternatives by incorporating predicted point cloud overlap scores and diffusion-based denoising of the estimated correlation matrices.

\noindent
\textbf{Multiview registration}
methods~\cite{govindu2004lie,torsello2011multiview,arie2012global,arrigoni2014robust,bernard2015solution,arrigoni2016spectral,maset2017practical,birdal2018bayesian,gojcic2020learning,yew2021learning,wang2023robust} reconstruct a complete scene from a collection of partially overlapping point clouds.
The first family of methods employs a multiview ICP to optimize camera poses and 3D correspondences~\cite{huber2003fully,mian2006three,fantoni2012accurate,birdal2017cad}. 
However, these methods often struggle with the increased correspondence estimation complexity.
To address this, some approaches focus solely on optimizing motion, using the point clouds to evaluate errors~\cite{theiler2015globally,zhou2016fast,bhattacharya2019efficient}.
However, such ICP-based methods are prone to inaccuracies in the pairwise poses that provide the starting point for the multi-view procedure.

Other modern methods take a different approach, using global cycle consistency to optimize poses starting from an initial set of pairwise maps. 
This so-called synchronization method is known for its efficiency~\cite{torsello2011multiview,arie2012global,arrigoni2014robust,theiler2015globally,bernard2015solution,zhou2016fast,maset2017practical,bhattacharya2019efficient,huang2019learning,birdal2019probabilistic,tejus2023rotation}. 
Global Structure-from-Motion~\cite{cui2015global} synchronizes observed relative motions and decomposes them into rotation, translation, and scale components. 
Our pipeline follows a similar global approach, with geometric optimization at its core, combined with recent advancements in deep learning to leverage the best of both worlds.

Recent methods~\cite{arrigoni2016spectral,choi2015robust,huang2017translation} employ an iteratively reweighted least-squares (IRLS) scheme to adaptively downweight noisy pairwise estimates. 
However, the iterative refinement of IRLS can become trapped in local minima and may fail to remove outlier edges. 
To tackle this challenge, recent learning-based advances~\cite{gojcic2020learning,huang2019learning,yew2021learning,wang2023robust} adopt a data-driven strategy to learn robust reweighting functions. 
While these approaches allow for end-to-end training, they often make design choices prioritizing ease of training over performance during inference.
In contrast, our paper takes a different approach.
Instead of prioritizing end-to-end trainability, which may not be directly relevant in real-world scenarios, we aim to design a framework estimating highly accurate multiway registration from a set of partially overlapping point clouds.
Our focus is on precision and reliability, guiding our design choices throughout the development of the framework.
\section{Pairwise and Multiway Registration}
\label{cha:method}

\noindent
\textbf{Problem Definition.}
Point cloud mosaicking from pairwise registrations can be formalized as a rigid transform averaging, recovering 3D orientations $\bm{R}_i \in \text{SO}(3)$ and positions $\bm{t}_i \in \mathbb{R}^3$ from a set of estimated relative pairwise motions $(\bm{R}_{ij}, \bm{t}_{ij})$, where $i, j \in [1, n]$ and $n \in \mathbb{N}$ is the number of partially overlapping point clouds ($i \neq j$). 
We can express the information as pose graph $\mathcal{G} = (\mathcal{V}, \mathcal{E})$, where each vertex $v \in \mathcal{V}$ represents a global pose, and each edge $(i, j) \in \mathcal{E}$ is the relative
motion of point clouds $\mathcal{P}_i$ and $\mathcal{P}_j$. 
The relative and global transforms
are related by constraints:
\begin{equation}
    \bm{R}_{ij} = \bm{R}_j \bm{R}_i^{\text{T}}, \quad \bm{t}_{ij} = \bm{R}_i (\bm{t}_j - \bm{t}_i), \quad \forall (i, j) \in \mathcal{E}.
    \label{eq:pairwise_constraint}
\end{equation}
Relative transforms are 
obtained from pairwise registration methods and are corrupted by noise and outliers. 
Thus, a solution that satisfies all constraints in Eq.~\ref{eq:pairwise_constraint} cannot be found. 
To circumvent this, transformation averaging seeks to recover global transforms with minimum consistency error. 

\vspace{0mm}
\noindent
\textbf{Pipeline Summary.}
The proposed pipeline (called Wednesday) performing pairwise and then multiway point cloud registration is depicted in Fig.~\ref{fig:teaser}. 
It begins by iterating through pairs of point clouds, where tentative 3D point correspondences are established and relative poses are estimated, as proposed in Section~\ref{sec:pairwise}. 
These pairs are utilized to construct a pose graph. 
To further refine the estimated poses, the pipeline adopts a decoupled approach. 
This involves first optimizing the global orientations as described in Section~\ref{sec:rot_avg}, then re-estimating the relative translations based on the global orientations (Section~\ref{sec:trans_reest}), and finally, optimizing the global positions as outlined in Section~\ref{sec:trans_opt}.
At last, diffusion-based optimization, detailed in Section~\ref{sec:diff_optimization}, is applied to further optimize the pose graph.

\subsection{Overlap and Diffusion-based Registration}
\label{sec:pairwise}

This section introduces the Overlap-aware, Diffusion-aided paIrwise registratioN (ODIN) methoD, enhancing SOTA frameworks like GeoTransformer~\cite{qin2023geotransformer} and PEAL~\cite{yu2023peal} by incorporating diffusion-based denoising and iteratively optimizing point cloud overlap within the attention learning.

\vspace{0mm} \noindent
\textbf{Feature extraction.}
ODIN begins by extracting features from individual points and superpoints (clusters of points), as in GeoTransformer~\cite{qin2023geotransformer}. 
Utilizing the KPConv-FPN backbone~\cite{thomas2019kpconv}, we downsample input point clouds to multi-level features $\mathbf{F}^{\hat{\mathcal{P}}} \in \mathbb{R}^{|\hat{\mathcal{P}}|\times d}$ and $\mathbf{F}^{\hat{\mathcal{Q}}} \in \mathbb{R}^{|\hat{\mathcal{Q}}|\times d}$. 
The sets of coarsest resolution points, treated as superpoints, are denoted as $\hat{\mathcal{P}}$ and $\hat{\mathcal{Q}}$, along with their associated features. 
Additionally, dense sets of correspondences, $\tilde{\mathcal{P}}$ and $\tilde{\mathcal{Q}}$, and their features $\mathbf{F}^{\tilde{\mathcal{P}}} \in \mathbb{R}^{|\tilde{\mathcal{P}}|\times d}$ and $\mathbf{F}^{\tilde{\mathcal{Q}}} \in \mathbb{R}^{|\tilde{\mathcal{Q}}|\times d}$ are computed at half the original resolution.
Each point in $\tilde{\mathcal{P}}$ is assigned to its nearest superpoint. 
The feature matrix associated with the points in $\mathcal{G}_i^{\tilde{\mathcal{P}}}$ is denoted as $\mathbf{F}_i^{\tilde{\mathcal{P}}} \subseteq \mathbf{F}^{\tilde{\mathcal{P}}}$.
Superpoints without assignments are removed. 
Patches $\{\mathcal{G}_i^{\hat{\mathcal{Q}}}\}$ and features $\{\mathbf{F}_i^{\hat{\mathcal{Q}}}\}$ for point cloud $\hat{\mathcal{Q}}$ are computed similarly.

\vspace{0mm} \noindent
\textbf{Geometric Self-Attention.}
Following~\cite{qin2023geotransformer}, we employ self-attention mechanisms within the superpoints \( \hat{\mathcal{P}} \) and \( \hat{\mathcal{Q}} \) to learn point-specific attention features. 
The self-attention is formalized as follows:
Given input feature matrix \( \bm{X} \in \mathbb{R}^{|\hat{\mathcal{P}}| \times d_t} \), we compute output feature matrix \( \bm{Z} \in \mathbb{R}^{|\hat{\mathcal{P}}| \times d_t} \). 
Each element in $\bm{Z}$ is a cumulative sum of the weighted, projected input features as:
    $\bm{Z}_i = \sum_{j=1}^{|\hat{\mathcal{P}}|} a_{i,j} (x_j \bm{W}^V)$,
where $a_{i,j}$ is the weight coefficient for the $i$th and $j$th superpoint, obtained via a row-wise softmax applied to the attention scores $e_{i,j}$. 
These scores are calculated as
    $e_{i,j} = ((\bm{x_i} \bm{W}^Q)(\bm{x_j} \bm{W}^K + \bm{r_{i,j}} \bm{W}^R)^T) /\sqrt{d_t}$,
where \( \bm{r_{i,j}} \in \mathbb{R}^{d_t} \) represents a vector embedding geometric structural information, capturing pairwise distances and angular relationships among points.
Projection matrices \( \bm{W}^Q \), \( \bm{W}^K \), \( \bm{W}^V \), and \( \bm{W}^R \in \mathbb{R}^{d_t \times d_t} \) correspond to queries, keys, values, and geometric structure embeddings, respectively.
The outcome of this process is matrices \( \bm{X}^{\hat{\mathcal{P}}} \) and \( \bm{X}^{\hat{\mathcal{Q}}} \), representing the learned attention features for superpoints in \( \hat{\mathcal{P}} \) and \( \hat{\mathcal{Q}} \).

We distinguish anchor and non-anchor superpoints, determined by an attention mask based on the overlap scores predicted later.
As the overlap is unknown in the first iteration, we use an identity mask (updated later), making all superpoints anchors.
Matrices \( \bm{X}^{\hat{\mathcal{P}}_A} \), \( \bm{X}^{\hat{\mathcal{Q}}_A} \) are features  for anchors, and \( \bm{X}^{\hat{\mathcal{P}}_N} \), \( \bm{X}^{\hat{\mathcal{Q}}_N} \) are for non-anchors.
At this point, the attention learning splits into two separate streams.
The steps discussed next are contributions of this paper. 

\vspace{0mm} \noindent
\textbf{A) Explicit One-Way and Cross-Attention.}
This stream is responsible for incorporating overlap information into attention learning. 
It starts with an explicit one-way attention module~\cite{yu2023peal} to learn the intra-frame correlations with anchor superpoints, which is critical to encode inter-frame geometric consistency. 
The module begins by differentiating anchor and non-anchor superpoints, working with their respective feature matrices $\bm{X}^{\hat{\mathcal{P}}_A}$ and $\bm{X}^{\hat{\mathcal{P}}_N}$. 
The attention features for non-anchor superpoints, denoted as \( \bm{Z}^{\hat{\mathcal{P}}_N} \) 
are computed by leveraging the attention features of the anchor superpoints as
$\bm{Z}^{\hat{\mathcal{P}}_N}_m = \sum_{n=1}^{|\hat{\mathcal{P}}_A|} \alpha_{m,n} \left( \bm{X}^{\hat{\mathcal{P}}_A}_n \bm{W}^V \right)$.
Here, $\alpha_{m,n}$ indicates the attention score, obtained through a row-wise softmax function, representing the feature correlation between non-anchor \( \bm{X}^{\hat{\mathcal{P}}_N} \) and anchor \( \bm{X}^{\hat{\mathcal{P}}_A} \) superpoints. 
The specific attention score $e_{m,n}$ is given by:
\begin{equation}
e_{m,n} = \frac{\left( \bm{X}^{\hat{\mathcal{P}}_N}_m \bm{W}^P_A \right) \left( \bm{X}^{\hat{\mathcal{P}}_A}_n \bm{W}^K \right)^T}{\sqrt{d_t}}.
\end{equation}
This is similarly applied to update attention features for \( \bm{X}^{\hat{\mathcal{Q}}_N} \), while features \( \bm{X}^{\hat{\mathcal{Q}}_A} \) and \( \bm{X}^{\hat{\mathcal{P}}_A} \) remain unchanged. 

\vspace{0mm} \noindent
\textbf{B) Cross-Attention Stream.}
Complementing the previous calculations, the framework employs another stream directly utilizing the self-attention embeddings, determined in feature extraction, to facilitate cross-attention learning~\cite{qin2023geotransformer}.

\vspace{0mm} \noindent
\textbf{Correlation Maps.} 
Each stream outputs a correlation map where the value in the $i$th row and $j$th column signifies the correlation of the $i$th superpoint in the first point cloud and the $j$th in the second. 
The final correlation map is obtained as a weighted sum of correlations from both streams. It is designed to evolve by being updated at the end of the pipeline, capturing overlap information by leveraging the predicted transformation from prior iterations.
The initial weight of one-way attention is set to zero. The upper stream will gradually gain more attention during training.
We also apply an attention mask to the weighted correlation matrix, in which high-confidence matches gain more attention. This significantly accelerates the training process.
Different from~\cite{yu2023peal}, our method can be trained from scratch and does not rely on the initial overlapping prediction.

\begin{figure*}[t]
\begin{center}
    \includegraphics[width=0.95\textwidth]{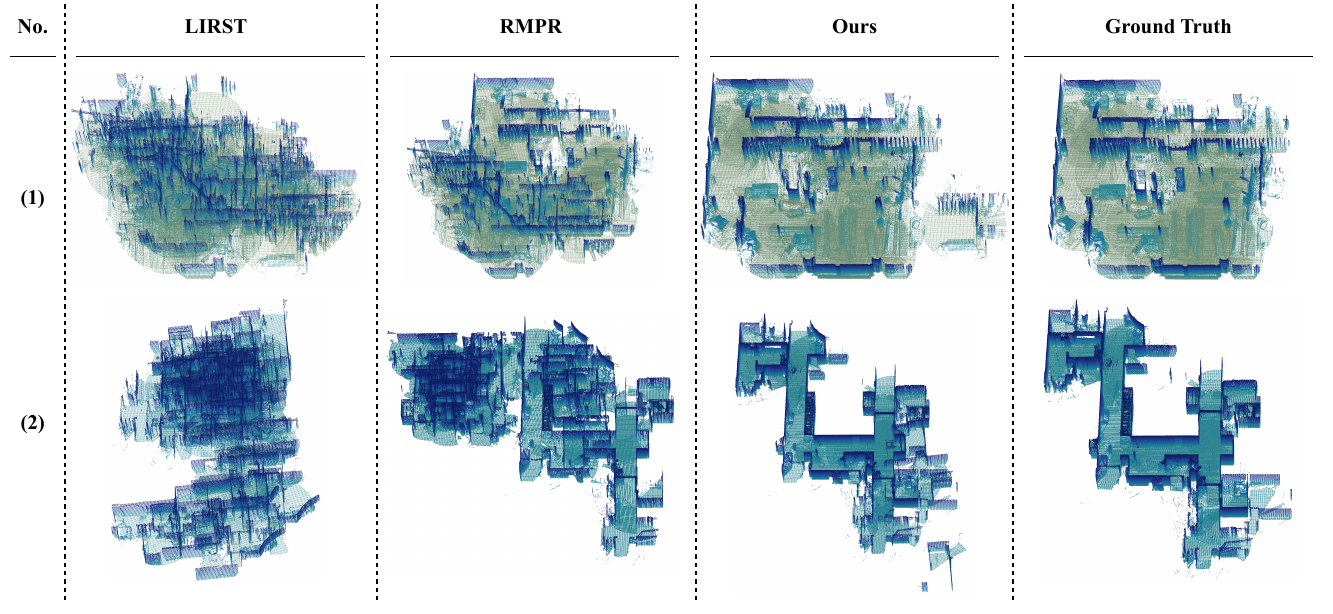}
\end{center}
    \vspace{-6mm}
    \caption[Qualitative Results]{\textbf{Multiway point cloud registration} results on two scenes from the challenging NSS dataset~\cite{nss2023} with the recent LIRST~\cite{yew2021learning} and RMPR~\cite{wang2023robust}  and the proposed methods (ceilings are not shown). Also, we visualize the provided ground truth. 
    We show results for LIRST and RMPR as they are the best-performing alternatives in Tab.~\ref{tab:NSS_mr}. Such results are common output of these methods on this dataset.
    }
    \label{fig:visualization}
    \vspace{-5mm}
\end{figure*}

\vspace{0mm} \noindent
\textbf{Correlation Denoising by Diffusion.}
This step focuses on improving the predicted correlations by reducing the effect of noise. 
We employ diffusion models, a type of probabilistic generative model that learns to transform a noisy sample \( h_K \sim \mathcal{N}(0, \mathbf{I}) \) into a clean one \( h_0 \). 
Each noisy $h_k$ is expressed as a linear mix of the source sample \( h_0 \) and the noise variable \( \epsilon \) as 
$h_k := \sqrt{\alpha_k}h_0 + \sqrt{1 - \alpha_k}\epsilon, \quad \epsilon \sim \mathcal{N}(0, 1)$.
Using sample \( h_0 \) and the forward diffusion-generated noisy samples \( \{h_k\}^K_{k=1} \), diffusion model \( g \) is optimized to approximate the reverse process.
Finally, the reverse step is recurrently performed to generate a high-quality sample \( h_0 \) from the noisy one $h_K$ using the trained model \( g \).

We assume that the correlation matrix generated previously is a noisy observation of the actual, unknown correlation. 
Thus, we use a noise reduction diffusion process, enhancing the subsequent matching procedure. 
Drawing inspiration from \cite{diffusionAutoencoder2023} and \cite{dhariwal2021diffusion}, we adopt a UNet architecture and the overlap-aware circle loss~\cite{qin2023geotransformer}. 
The loss on superpoint patch \( G_{i}^{P} \) (\ie, a continuous representation of the underlying local surface) is defined as follows:
\begin{eqnarray*}
    \footnotesize
    L_{oc}^{P} = \frac{1}{|A|} \sum_{G_{i}^{P} \in A}  \\  \log\left(1 + \sum_{G_{j}^{Q} \in \epsilon_{p}} e^{\beta_{i, j}^{p}\lambda_{i}^{j} \left(d_{i}^{j} - \Delta_{p}\right)} - \sum_{G_{k}^{Q} \in \epsilon_{n}} e^{\beta_{i, k}^{n} \left(\Delta_{n} - d_{i}^{k}\right)}\right),
\end{eqnarray*}
where \(d_{i}^{j} = \|\hat{h}_{i}^{P} - \hat{h}_{j}^{Q}\|_{2}\) is the distance in the feature space,
\(\lambda_{i}^{j} = ( o_{i}^{j} )^{1/2}\) and \(o_{i}^{j}\) represents the overlap ratio between \(G_{i}^{P}\) and \(G_{j}^{Q}\).
The positive and negative weights are computed for each sample individually as \(\beta_{p}^{i,j} = \gamma(d_{i}^{j} - \Delta_{p})\) and \(\beta_{n}^{i,k} = \gamma(\Delta_{n} - d_{i}^{k})\). 
The loss \(L_{oc}^{Q}\) on $Q$ is calculated similarly. 
The overall loss is \(L_{oc} = (L_{oc}^{P} + L_{oc}^{Q}) / 2\).
The operational model is conditioned on the features of the superpoints. These features undergo iterative updates during each step of the denoising process. 
The overlap-aware circle loss is employed as a strategic optimization mechanism to refine and optimize these features. 
The foundational ground truth correlation matrix is the normalized matrix of overlapping ratios, where each constituent element represents the overlapping proportion of respective superpoint patches. 

\begin{table*}[t]
    \footnotesize
    \begin{center}
    \resizebox{\linewidth}{!}{
    \begin{tabular}{l|ccc|ccc|ccc|ccc}
        \toprule
        \multirow{2}{*}{Method} & \multicolumn{3}{c|}{NSS~\cite{nss2023}} & \multicolumn{3}{c|}{3DMatch~\cite{zeng20173dmatch}} & \multicolumn{3}{c|}{3DLoMatch~\cite{huang2021predator}} & \multicolumn{3}{c}{KITTI~\cite{huang2021predator}} \\
        & RR (\%)$\uparrow$ & RRE ($^{\circ}$)$\downarrow$ & RTE (m)$\downarrow$ & RR (\%)$\uparrow$ & RRE ($^{\circ}$)$\downarrow$ & RTE (m)$\downarrow$ & RR (\%)$\uparrow$ & RRE ($^{\circ}$)$\downarrow$ & RTE (m)$\downarrow$ & RR (\%)$\uparrow$ & RRE ($^{\circ}$)$\downarrow$ & RTE (cm)$\downarrow$ \\
	    \midrule
	  FPFH~\cite{rusu2009FPFH} & 11.70 & 45.32 & 2.23 & 0.851 & -- & -- & -- & -- & -- & -- & -- & -- \\
        FCGF~\cite{Choy2019FCGF} & 24.43 & 39.89 & 2.04 & -- & -- & -- & 0.401 & -- & -- & 96.0 & 0.30 & 9.5 \\
        D3Feat~\cite{bai2020d3feat} & 22.73 & 33.09 & 2.26 & 0.816 & -- & -- & 0.372 & -- & -- & \textbf{99.8} & 0.30 & 7.2 \\
        RegTR~\cite{yew2022regtr} & -- & -- & -- & 0.919 & 5.31 & 0.170 & 0.646 & 23.05 & 0.644 & -- & -- & -- \\
        Predator~\cite{huang2021predator} & 64.97 & 13.52 & 0.65 & 0.893 & 6.80 & 0.202 & 0.604 & 30.07 & 0.762 & \textbf{99.8} & 0.27 & 6.8 \\
        GeoTr.~\cite{qin2023geotransformer} & 39.07 & 22.93 & 0.99 & 0.925 & 7.04 & 0.194 & 0.741 & 23.15 & 0.583 & \textbf{99.8} & 0.24 & 6.8 \\
        PEAL~\cite{yu2023peal} & 58.72 & 15.78 & 0.71 & 0.941 & 4.23 & 0.152 & 0.788 & 15.79 & 0.485 &  \textbf{99.8} & 0.23 & 6.8 \\
        \textbf{ODIN} & \textbf{69.73} & \textbf{11.96} & \textbf{0.54} & \textbf{0.958} & \textbf{3.15} & \textbf{0.108} & \textbf{0.812} & \textbf{12.61} & \textbf{0.402} & \textbf{99.8} & \textbf{0.14} & \textbf{3.6} \\
    	\bottomrule
    \end{tabular}
    }
    \end{center}
    \vspace{-5mm}
\caption{\textbf{Pairwise point cloud registration} on the NSS~\cite{nss2023}, 3DMatch~\cite{zeng20173dmatch}, 3DLoMatch~\cite{huang2021predator} and KITTI~\cite{geiger2012we}  datasets. The reported metrics are the Registration Recall (RR), which measures the fraction of successfully registered pairs;
the Relative Rotation Error (RRE); and the Relative Translation Error (RTE). The best results are in \textbf{bold}. }
\label{tab:NSS}
\vspace{-5mm}
\end{table*}

\vspace{0mm} \noindent
\textbf{Point Matching Module.}
With the correlation matrix denoised, the next step establishes superpoint correspondences and extracts 3D point correspondences through the Point Matching Module. 
This step is similar to what is done in~\cite{qin2023geotransformer}.

For each identified superpoint match \( \hat{C}_i = (\hat{\mathcal{P}}_{x_i}, \hat{\mathcal{Q}}_{y_i}) \), we employ an optimal transport layer to ascertain dense point correspondences between \( G^P_{x_i} \) and \( G^Q_{y_i} \). 
The process begins by constructing a cost matrix \( \bm{C}_i \) as follows:
$\bm{C}_i = \bm{F}^P_{x_i} (\bm{F}^Q_{y_i})^T / \sqrt{d}$,
where \( n_i = |G^P_{x_i}| \) and \( m_i = |G^Q_{y_i}| \). 
Cost matrix \( \bm{C}_i \) is augmented by appending a new row and column filled with a learnable dustbin parameter \( \alpha \). 
We convert \( \bm{C}_i \) into a soft assignment matrix \( \bm{Z}_i \) with the Sinkhorn algorithm, which serves as the confidence matrix for candidate matches.
Point correspondences are determined through mutual top-$k$ selection, identifying matches as those ranking among the top $k$ entries in their respective rows and columns:
    $C_i = \left\{ (G^P_{x_i}(x_j), G^Q_{y_i}(y_j)) | (x_j, y_j) \in \text{mutual\_top}_{k} \left( \bm{Z}_i^{x, y} \right) \right\} $.
The correspondences from each superpoint match are combined to construct dense correspondences as $C = \bigcup_{i=1}^{N_c} C_i$.
They are then used to regress the rigid pose parameters.

Subsequently, the computed pose is employed to assess the overlap of point clouds via a nearest neighbor search in the Euclidean space as done in~\cite{yu2023peal}.
This overlap score is considered a 3D overlap prior.
In contrast to prior work, we reintegrate this score into the attention learning module, which restarts the learning process with updated attention masks. 
This iterative procedure is repeated $I$ times, progressively learning the overlap-aware attention scores, thereby enabling accurate matching of dense 3D point pairs.

Given the estimated relative transforms, we construct pose graph $\mathcal{G} = (\mathcal{V}, \mathcal{E})$ with $\mathcal{V}$ representing the global poses of individual point clouds and $\mathcal{E}$ denoting the relative transforms between them, estimated previously.
We will now discuss the proposed multiway algorithm: Wednesday. 

\subsection{Global Rotation Averaging}
\label{sec:rot_avg}

The goal of rotation averaging is to deduce the global orientations decoupled from the positions given pairwise rotations. 
We employ the method of \cite{chatterjee2013efficient} to obtain global point cloud orientations, finding it exceptionally efficient for our problem. 
In brief, this method performs a two-step procedure. 
First, it employs an L1 optimization to yield coarse rotation estimates robust to outliers. 
Next, it utilizes an iteratively re-weighted least-squares approach to refine these initial estimates and obtain accurate global rotations.

Selecting the somewhat older method of \cite{chatterjee2013efficient} might seem unconventional when aiming for a pipeline that delivers state-of-the-art accuracy. 
However, this approach proved to be the most applicable out-of-the-box solution for our problem.
We explored several recent learning-based alternatives, including NeuRoRa \cite{purkait2020neurora}, PoGO-Net \cite{li2021pogo}, MSP~\cite{yang2021end}, and DMF-Net \cite{tejus2023rotation}. 
Although NeuRoRa and MSP demonstrate commendable accuracy in their experiments, we could not reproduce these results, even after retraining the models. 
PoGO-Net lacks a public implementation. 
DMF-Net does not yield better results than \cite{chatterjee2013efficient} according to their own experiments, and its optimization process is particularly time-consuming, especially when compared to \cite{chatterjee2013efficient} that runs only for a few seconds in practice.

\begin{table*}[h]
    \footnotesize
    \begin{center}
    \resizebox{0.85\linewidth}{!}{
    \begin{tabular}{l|cc|l|cc|cc|cc}
        \toprule
        \multirow{2}{*}{Method} & \multicolumn{2}{c|}{NSS} & & \multicolumn{2}{c|}{3DMatch} & \multicolumn{2}{c|}{3DLoMatch} & \multicolumn{2}{c}{KITTI} \\ 
        & RE ($^{\circ}$)$\downarrow$ & TE (m)$\downarrow$ & & RE ($^{\circ}$)$\downarrow$ & TE (cm)$\downarrow$ & RE ($^{\circ}$)$\downarrow$ & TE (cm)$\downarrow$& RE ($^{\circ}$)$\downarrow$ & TE (cm)$\downarrow$ \\
        \midrule
        Predator & 13.43 & 0.65 & PEAL & 4.72 & 15.8 & 16.03 & 50.2 & 9.46 & 11.85 \\
        {+ } Open3d~\cite{choi2015robust} & 12.76 & 0.64 & + Open3d & 4.72 & 15.8 & 14.23 & 45.1 & 6.21 & \phantom{1}7.72 \\
        {+ } DeepMapping2~\cite{chen2023deepmapping2} & 11.54 & 0.64 & + DeepM.\ & 4.23 & 14.5 & 13.25 & 39.4 & 3.34 & \phantom{1}6.04 \\
        {+ } LMPR~\cite{gojcic2020learning} & 11.35 & 0.62 & + LMPR & 3.98 & 12.6 & 13.07 & 37.3 & 6.79 & \phantom{1}7.89 \\
        {+ } LIRTS~\cite{yew2021learning} & 11.42 & 0.61 & + LIRTS & 3.95 & 12.0 & 11.52 & 36.0 & 5.17 & \phantom{1}6.94  \\
        {+ } RMPR~\cite{wang2023robust} & 10.87 & 0.62 & + RMPR & 3.57 & 11.6 & 10.18 & 34.4 & 4.69 & \phantom{1}6.38 \\
        {+ } \textbf{Wednesday} & \phantom{1}2.24 & 0.51 & + \textbf{Wednesday} & 2.58 & \phantom{1}9.4 & \phantom{1}7.21 & 29.1 & 2.52 & \phantom{1}5.92 \\
        \textbf{ODIN} + \textbf{Wednesday} & \phantom{1}\textbf{2.01} & \textbf{0.42} & & \textbf{2.32} & \phantom{1}\textbf{8.4} & \phantom{1}\textbf{6.44} & \textbf{26.5} & \textbf{2.18} & \phantom{1}\textbf{4.76} \\
        \bottomrule
    \end{tabular}
    }
    \end{center}
    \vspace{-5mm}
\caption{\textbf{Multiway point cloud registration} on the NSS~\cite{nss2023}, 3DMatch~\cite{zeng20173dmatch}, 3DLoMatch~\cite{huang2021predator} and KITTI~\cite{geiger2012we}  datasets. 
The reported metrics are the average rotation (RE) and translation errors (TE). 
For each dataset, we choose the best-performing pairwise estimator from the baselines (see Table~\ref{tab:NSS}). 
We run Predator~\cite{huang2021predator} on NSS and PEAL~\cite{yu2023peal} on the other datasets. The best results are in \textbf{bold}. }
\vspace{-5mm}
\label{tab:NSS_mr}
\end{table*}

\subsection{Optimal Robust Translation Re-estimation}
\label{sec:trans_reest}

The next phase in our pipeline is the re-estimation of relative translations $\bm{t}_{ij}$, using the estimated global point cloud orientations $\{ \bm{R}_i \}_{i \in [1, n]}$. This step is crucial as the initial relative translations are computed alongside rotations, posing a more complex problem than the robust estimation of Euclidean translations with known orientations. This phase is key to achieving better translation initialization for the subsequent position averaging stage.

Given rotations $\bm{R}_i$ and $\bm{R}_j$, and point correspondence $(\bm{X}_i, \bm{X}_j) \in \mathcal{X}_{ij}$ in frames $i$ and $j$, we have constraint
    $\bm{R}_j \bm{X}_j = \bm{R}_i \bm{X}_i + \bm{t}_{ij}$,
where $\bm{t}_{ij} \in \mathbb{R}^3$ is the unknown relative translation. With known $\bm{R}_i$ and $\bm{R}_j$, we derive equation
\begin{equation}
\bm{t}_{ij} = \bm{R}_j \bm{X}_j - \bm{R}_i \bm{X}_i
\label{eq:translation_reest}
\end{equation}
to calculate the updated translation from a single correspondence.
To address the outlier correspondences, methods like RANSAC~\cite{fischler1981random} or L1 optimization~\cite{hartley2011l1} could be applied. 
However, our problem allows for solving the maximum consensus problem -- finding the model with the highest number of inliers -- in a globally optimal fashion.

Assuming an inlier-outlier threshold $\epsilon \in \mathbb{R}^+$, our objective is to find optimal translation 
\begin{equation*}
    \bm{t}_{ij}^* = \arg \max_{\bm{t}_{ij}} \sum_{(\bm{X}_i, \bm{X}_j) \in \mathcal{X}_{ij}} \iverson{ \left | \bm{t}_{ij} - \bm{R}_j \bm{X}_j + \bm{R}_i \bm{X}_i \right|_2 < \epsilon },
\end{equation*}
where $\iverson{\cdot}$ is the Iverson bracket that equals 1 if the condition inside holds and 0 otherwise. 
In our case, this translates to finding the maximum overlap of 3D spheres.

For each translation estimate $\hat{\bm{t}}_{ij}$ coming from a 3D correspondence via Eq.~\ref{eq:translation_reest}, the points that lead to translations falling inside a sphere with radius $\epsilon$ centered on $\hat{\bm{t}}_{ij}$ are the inliers. 
Therefore, finding the 3D subspace with the highest sphere density yields the solution with the maximum inliers.
This is a similar process as proposed for 1D problems in \cite{barath2021image}. 

Analytically solving the maximum sphere overlap problem is complex. 
Instead, we propose a fast ``branch-and-bound''-like numerical approach that achieves global optimum. 
The process starts by iterating through all correspondences $(\bm{X}_i^k, \bm{X}_j^k) \in \mathcal{X}_{ij}$, calculating the implied translations $\hat{\bm{t}}_{ij}^k$. 
We then create a uniform 3D grid where each cell (\ie, a 3D box) counts the number of intersecting spheres. 
The cell, with a size arbitrarily set to $\epsilon$ for this stage, allows efficient calculation of box-sphere intersections by intersecting the boundary planes of a cell with the spheres. 
The intersections can fall into three categories: 
(i) The sphere does not intersect any boundary planes, and the distance functions indicate it is outside the box, 
(ii) the sphere intersects one or more planes, or 
(iii) the sphere is entirely within the box as indicated by the distance functions. 
We repeatedly zoom into the cells with the highest sphere density, re-initializing a uniform grid on these selected cells.
The procedure stops at the zoom level, where the selected cells fall inside the affected spheres without intersecting their boundaries. 
As we may end up with multiple cell candidates, we choose the one with the highest inlier number. 
Finally, the translation is estimated by least squares fitting on all inliers.

It is important to note that with a sparse grid and proper hashing functions, this procedure is $\mathcal{O}(|\mathcal{X}|)$, scaling linearly with the number of correspondences. 
In practice, only the first step checks all correspondences, with subsequent steps focusing on a significantly smaller set of candidates.

\subsection{Translation Optimization}
\label{sec:trans_opt}

Given the estimated global orientations and refined relative translations, the next step is estimating the global positions of the point clouds. 
To do so, we employ a Levenberg-Marquardt numerical optimization implemented in the Ceres library, minimizing pairwise constraints $\bm{t}_{ij} = \bm{R}_i (\bm{t}_j - \bm{t}_i)$. 
We use the truncated soft-L1 robust loss.

\subsection{Pose Graph Optimization with Diffusion}
\label{sec:diff_optimization}

Given the global poses estimated in the previous sections, the last step of the algorithm is a joint optimization of positions and orientations. 
Inspired by \cite{wang2023posediffusion}, we design a denoising network to model the conditional probability \( p(\bm{R}, \bm{t} \; | \; \mathcal{K}) \) of the samples ($\bm{R}, \bm{t}$) given the set of input point clouds $\mathcal{K}$. 
Probability $p(\bm{R}, \bm{t} \; | \; \mathcal{K})$ is first estimated by training a diffusion model \( \mathcal{D}_{\theta} \) on point clouds with ground truth poses from a training set. 
At inference time, for a new set of point clouds \( \mathcal{K} \), we sample \( p(\bm{R}, \bm{t} \; | \; \mathcal{K}) \) to estimate the corresponding global pose \( \bm{R}, \bm{t} \).

The denoising process is conditioned on the input point cloud set $\mathcal{K}$, as \( p_{\theta}(\bm{R}_{t -1}, \bm{t}_{t - 1} \; | \; \bm{R}_{t}, \bm{t}_{t}, \mathcal{K}) = \)
\begin{eqnarray*}
    \footnotesize
    \mathcal{N} \left(\bm{R}_{t-1}, \bm{t}_{t - 1}; \sqrt{1 - \beta_t} \mathcal{D}_{\theta}(\bm{R}_{t}, \bm{t}_{t}, t, \mathcal{K}), (1 - \beta_t)\mathcal{K} \right).
\end{eqnarray*}
Denoiser \(\mathcal{D}_{\theta}\) is implemented as a Transformer, which accepts a sequence of noisy poses \(\bm{R}_t^i, \bm{t}_t^i\), diffusion time \(t\), and feature embeddings \(\psi(P^i) \in \mathbb{R}^{D_{\psi}}\) of the input point cloud \(\mathcal{K}^i\). The denoiser outputs the tuple of corresponding denoised pose parameters \(\mu_{t-1} = (\mu_{t-1}^i)_{i=1}^N\). 
The feature embeddings come from a pretrained KPConv.

\noindent
At train time, \(\mathcal{D}_{\theta}\) is supervised by denoising loss $\mathcal{L}_{\text{diff}} = $
\begin{equation*}
    \small
    \mathbb{E}^{t \sim \mathcal{U}[1,T]}_{\bm{R}_{t}, \bm{t}_{t} \sim p(\bm{R}_{t}, \bm{t}_{t} | \bm{R}_{0}, \bm{t}_{0}, \mathcal{K})} \left\| \mathcal{D}_{\theta}(\bm{R}_{t}, \bm{t}_{t}, t, P) - (\bm{R}_{t}. \bm{t}_{t}) \right\|^2. 
\end{equation*}
The relative poses from the pairwise registration constrain the whole pose graph.
The error implied by an edge is
$\epsilon(\bm{R}_{ij}, \bm{t}_{ij}) = \sqrt{ \frac{1}{|\mathcal{C}|} \sum_{(p, q) \in \mathcal{C_{ij}}} \| \bm{R} p_i + \bm{t} - q_j \|_2^2 }$,
where $\mathcal{C}_{ij}$ are the point correspondences in the point clouds.
The additional loss we designed to be minimized is $\mathcal{L} = \sum_{(i, j) \in \mathcal{E}} \min(\epsilon(\bm{R}_{ij}, \bm{t}_{ij}), \gamma)$, where $\mathcal{E}$ is the pose graph edges  and $\gamma$ is a threshold parameter.

The main differences compared to the original method in \cite{wang2023posediffusion} are the edge loss, measuring global pose consistency, and the employed denoising network architecture.

\section{Experiments}
\label{cha:experiments}

\textbf{Datasets.} To evaluate the proposed algorithms both on pairwise and multiway registration tasks, we use the \textit{3DMatch}~\cite{zeng20173dmatch}, \textit{3DLoMatch}~\cite{huang2021predator}, \textit{KITTI}~\cite{geiger2012we}, and \textit{NSS}~\cite{nss2023} datasets. 
The \textit{3DMatch}~\cite{zeng20173dmatch} dataset contains 62 indoor scenes, with 46 used for training, 8 for validation, and 8 for testing. We use the training data preprocessed by Huang et al.~\cite{huang2021predator} and evaluate on both \textit{3DMatch} and \textit{3DLoMatch}~\cite{huang2021predator} protocols. The point cloud pairs in \textit{3DMatch} have more than 30\% overlap, whereas those in \textit{3DLoMatch} have a low overlap of 10\% - 30\%.
The \textit{KITTI} odometry dataset~\cite{geiger2012we} contains 11 sequences of LiDAR-scanned outdoor driving scenarios. We follow~\cite{bai2020d3feat, Choy2019FCGF, huang2021predator, qin2022geometric} and split it into train/val/test sets as follows: sequences 0-5 for training, 6-7 for validation and 8-10 for testing. As in~\cite{bai2020d3feat, Choy2019FCGF, huang2021predator, qin2022geometric}, we refine the provided ground truth poses using ICP~\cite{besl1992method} and only use point cloud pairs that are captured within 10m range of each other. 
For multiway KITTI, we followed~\cite{chen2023deepmapping2} and decreased the frame rate 20 times to avoid saturated results.
The \textit{NSS} dataset represents 6 large-scale construction sites and their rescans over time. The data depicts the interior layout construction from creating walls, to adding pipes and air-ducts, and to machinery moving around. It contains spatial and spatiotemporal pairs and has annotations for both pairwise and multiway registration.  

\vspace{0mm}
\noindent
\textbf{Metrics.}
For evaluating pairwise methods, we follow prior work~\cite{huang2021predator,qin2022geometric,yew2022regtr,jin2023q}. 
We compute the Registration Recall (RR), which measures the fraction of successfully registered pairs;
relative rotation (RRE); and relative translation errors (RTE).
We calculate the average values over all valid pairs and scenes.
For multiway registration, we calculate the average rotation (RE; in degrees) and position (TE) errors given the ground truth pose parameters.

\noindent
\textbf{Pairwise Point Cloud Registration.}
The results of the standard setting on the \textbf{NSS} dataset are in Table~\ref{tab:NSS} (1st col.).
ODIN substantially improves compared with the state-of-the-art methods.
We improve upon the recent PEAL~\cite{yu2023peal} by a margin of $11\%$ in terms of recall while reducing the trans.\ error by ${\approx}0.2$m and the rot.\ one by ${\approx}4^\circ$. 
Interestingly, the second best method on this dataset is Predator~\cite{huang2021predator}, which also significantly lags behind the proposed ODIN. 

The results on \textbf{3DMatch} are in Table~\ref{tab:NSS} (2nd).
While all methods are accurate, ODIN still manages to reduce the rotation by $1^\circ$ and translation errors by $0.05$ meters compared to the second best algorithm, PEAL. 
This improvement is also reflected in the recall, which is improved by $1.7\%$. 

The results on \textbf{3DLoMatch} are shown in Table~\ref{tab:NSS} (3rd).
On this dataset, we achieve significant improvements compared with other baselines. 
We improve upon the second most accurate method (GeoTransformer) by reducing the average rotation error by $10^\circ$ and the translation error by $0.18$ meters. 
This accuracy improvement pushes our recall score up by $7.1\%$ compared to that of GeoTransformer. 

The pairwise registration results on the \textbf{KITTI} dataset are in Table~\ref{tab:NSS} (4th).
While all methods perform very accurately, mainly due to the small baselines between subsequent frames, the proposed ODIN is the best in all metrics. 
Notably, it almost halves the rotation and position errors of the second most accurate method (PEAL). 
This clearly shows the advantages of the proposed two-stream architecture with attention masking and diffusion-based denoising.

\noindent
\textbf{Multiway Point Cloud Registration.}
We compare Wednesday to \cite{choi2015robust} implemented in Open3d, DeepMapping2~\cite{chen2023deepmapping2}, LMPR~\cite{gojcic2020learning}, LIRTS~\cite{yew2021learning}, and RMPR~\cite{wang2023robust}. 
We train all learning-based methods on the training set of each dataset.
To perform pairwise registration, we select the best-performing baseline method on each scene.
Thus, we run Predator~\cite{huang2021predator} on NSS and PEAL~\cite{yu2023peal} on the other datasets. 
Also, we show the results with the proposed ODIN. 

The results on all datasets are reported in Table.~\ref{tab:NSS_mr}.
The proposed Wednesday consistently improves upon \textit{all} state-of-the-art algorithms, often by a substantial margin. 
For example, the rotations errors on NSS are reduced to 20\%. 
The position errors on 3DMatch and 3DLoMatch are reduced by 2.2 and 5.3 meters, respectively. 
The rotation errors on KITTI are halved.
Using ODIN as a pairwise estimator further reduces the registration errors. 

We show results of the proposed method and \cite{yew2021learning,wang2023robust}
in Fig.~\ref{fig:visualization}. 
We chose \cite{yew2021learning,wang2023robust} as they are the best-performing alternatives in Table~\ref{tab:NSS_mr}, still, 
they fail entirely.
While the proposed method also has inaccuracies compared to the ground truth, it provides significantly better registrations.

\noindent
\textbf{Ablation Studies.}
The ablation study of pairwise registration on the {3DLoMatch} dataset is in Table~\ref{tab:ab_3DLoMatch}.
We tested ODIN without the proposed procedure of reintegrating the overlap score predictions from previous iterations, and without diffusion-based correlation matrix denoising. 
It is evident that both advancements have a clear and individual impact on the increased accuracy. 

We performed a similar ablation study on Wednesday on the NSS dataset, using ODIN to initialize the pairwise relative poses.
The results are in Table~\ref{tab:ab_NSS_mr}.
The diffusion-based optimization on its own leads to the highest rotations and second highest translation errors, justifying the need for the rest of the proposed pipeline.
This is expected as diffusion essentially aims at noise reduction, while the pose graph is not only noisy but contains outliers, necessitating robust estimation. 
Rotation and translation averaging leads to better results than the diffusion-based process.
It is further improved by re-estimating the translations.
The best results are obtained when all proposed components are employed. 

In Table~\ref{tab:transreest_comparison}, we demonstrate that the proposed globally optimal translation re-estimation outperforms running exhaustive RANSAC -- estimating the translation from each correspondence and selecting the one with the most inliers. 
The proposed approach is $2.4$ times faster while, as expected, it finds more inliers. 
Please note that this is the run-time on a single point cloud pair, thus the difference is more significant on the entire set of input point clouds. 

\begin{table}[t]
    \footnotesize
    \begin{center}
    \resizebox{0.80\columnwidth}{!}{
    \begin{tabular}{l|ccc}
        \toprule
        Method & RR (\%)$\uparrow$ & RRE ($^{\circ}$)$\downarrow$ & RTE (m)$\downarrow$ \\
        \midrule
        w/o Overlap and Diffusion & 0.741 & 23.15 & 0.583 \\
        w/o Overlap & 0.791 & 14.76 & 0.442 \\
        w/o Diffusion & 0.803 & 13.57 & 0.419 \\
        \textbf{ODIN} & \textbf{0.812} & \textbf{12.61} & \textbf{0.402} \\
        \bottomrule
    \end{tabular}
    }
    \end{center}
    \vspace{-5mm}
\caption{Pairwise registration recall (RR), rotation (RRE) and translation errors (RTE) of the proposed ODIN on the 3DLoMatch dataset without overlap scores or denoising the correlation matrix.}
\vspace{-10pt}
\label{tab:ab_3DLoMatch}
\end{table}

\begin{table}[t]
    \footnotesize
    \begin{center}
    \resizebox{\columnwidth}{!}{
    \begin{tabular}{l|ccccc}
        \toprule
        & D & R+TA & R+TR+TA & R+TA+D & R+TR+TA+D \\
        \midrule
        RE ($^{\circ}$)$\downarrow$ & 5.21 & 4.05 & 4.05 & 2.06 & \textbf{2.01} \\
        TE (m)$\downarrow$ & 0.51 & 0.53 & 0.48 & 0.47 & \textbf{0.42} \\
        \bottomrule
    \end{tabular}
    }
    \end{center}
    \vspace{-5mm}
\caption{Multiway registration average rotation and translation errors on the NSS dataset with combinations of the proposed components: (R) rotation and (TA) translation averaging, (TR) translation re-estimation, and (D) diffusion-based pose optimization. }
\vspace{-10pt}
\label{tab:ab_NSS_mr}
\end{table}

\begin{table}[t]
    \footnotesize
    \begin{center}
    \begin{tabular}{l|cc}
        \toprule
        Method & Runtime (s) $\downarrow$ & Inlier Number $\uparrow$ \\
        \midrule
        Exhaustive RANSAC & 0.12 & 18.03 \\
        Proposed (Sec.~\ref{sec:trans_reest}) & \textbf{0.05} & \textbf{18.11} \\
        \bottomrule
    \end{tabular}
    \end{center}
    \vspace{-5mm}
\caption{Average runtime (secs) and inlier number of exhaustive RANSAC and the proposed optimal estimator on the NSS dataset.}
\vspace{-12pt}
\label{tab:transreest_comparison}
\end{table}

\section{Conclusion}
\label{cha:conclusion}

We present Wednesday, a novel framework for multiway point cloud mosaicking, or else, aligning a collection of point clouds into a unified coordinate system. 
It starts with a new pairwise registration method (ODIN) which delivers significantly more accurate results compared to state-of-the-art ones. 
The pipeline proceeds with rotation and translation averaging to establish the global pose of each point cloud.
We also incorporate a globally optimal robust translation re-estimation algorithm to ensure the precision of pairwise translations after acquiring global orientations.
Finally, a diffusion-based optimization approach finalizes the output poses.
The pipeline leads to substantial improvement over the state-of-the-art algorithms, exemplified by an 80\% reduction in rotation error on the NSS dataset. 
The consistent and significant improvements on all tested large-scale datasets position the proposed algorithm as the new benchmark in both pairwise and multiway point cloud registrations.

{
    \small
    \bibliographystyle{ieeenat_fullname}
    \bibliography{main}

\begin{thebibliography}{86}
\providecommand{\natexlab}[1]{#1}
\providecommand{\url}[1]{\texttt{#1}}
\expandafter\ifx\csname urlstyle\endcsname\relax
  \providecommand{\doi}[1]{doi: #1}\else
  \providecommand{\doi}{doi: \begingroup \urlstyle{rm}\Url}\fi

\bibitem[Aiger et~al.(2008)Aiger, Mitra, and Cohen-Or]{aiger20084}
Dror Aiger, Niloy~J Mitra, and Daniel Cohen-Or.
\newblock 4-points congruent sets for robust pairwise surface registration.
\newblock In \emph{ACM SIGGRAPH 2008 papers}, pages 1--10. 2008.

\bibitem[Arie-Nachimson et~al.(2012)Arie-Nachimson, Kovalsky,
  Kemelmacher-Shlizerman, Singer, and Basri]{arie2012global}
Mica Arie-Nachimson, Shahar~Z Kovalsky, Ira Kemelmacher-Shlizerman, Amit
  Singer, and Ronen Basri.
\newblock Global motion estimation from point matches.
\newblock In \emph{2012 Second international conference on 3D imaging,
  modeling, processing, visualization \& transmission}, pages 81--88. IEEE,
  2012.

\bibitem[Arrigoni et~al.(2014)Arrigoni, Magri, Rossi, Fragneto, and
  Fusiello]{arrigoni2014robust}
Federica Arrigoni, Luca Magri, Beatrice Rossi, Pasqualina Fragneto, and Andrea
  Fusiello.
\newblock Robust absolute rotation estimation via low-rank and sparse matrix
  decomposition.
\newblock In \emph{2014 2nd International Conference on 3D Vision}, pages
  491--498. IEEE, 2014.

\bibitem[Arrigoni et~al.(2016)Arrigoni, Rossi, and
  Fusiello]{arrigoni2016spectral}
Federica Arrigoni, Beatrice Rossi, and Andrea Fusiello.
\newblock Spectral synchronization of multiple views in se (3).
\newblock \emph{SIAM Journal on Imaging Sciences}, 9\penalty0 (4):\penalty0
  1963--1990, 2016.

\bibitem[Author et~al.(2023)Author, Contributor, and
  Other]{diffusionAutoencoder2023}
First Author, Second Contributor, and Third Other.
\newblock Diffusion autoencoders: Toward a meaningful and decodable
  representation.
\newblock In \emph{Proceedings of the International Conference on Learning
  Representations}, pages 123--134. ICLR, 2023.

\bibitem[Bai et~al.(2020)Bai, Luo, Zhou, Fu, Quan, and Tai]{bai2020d3feat}
Xuyang Bai, Zixin Luo, Lei Zhou, Hongbo Fu, Long Quan, and Chiew-Lan Tai.
\newblock D3feat: Joint learning of dense detection and description of 3d local
  features.
\newblock In \emph{CVPR}, 2020.

\bibitem[Barath and Matas(2018)]{barath2018graph}
Daniel Barath and Ji{\v{r}}{\'\i} Matas.
\newblock Graph-cut ransac.
\newblock In \emph{Proceedings of the IEEE conference on computer vision and
  pattern recognition}, pages 6733--6741, 2018.

\bibitem[Barath et~al.(2021)Barath, Ding, Kukelova, and
  Larsson]{barath2021image}
Daniel Barath, Yaqing Ding, Zuzana Kukelova, and Viktor Larsson.
\newblock Image stitching with locally shared rotation axis.
\newblock In \emph{2021 International Conference on 3D Vision (3DV)}, pages
  1382--1391. IEEE, 2021.

\bibitem[Bernard et~al.(2015)Bernard, Thunberg, Gemmar, Hertel, Husch, and
  Goncalves]{bernard2015solution}
Florian Bernard, Johan Thunberg, Peter Gemmar, Frank Hertel, Andreas Husch, and
  Jorge Goncalves.
\newblock A solution for multi-alignment by transformation synchronisation.
\newblock In \emph{Proceedings of the IEEE conference on computer vision and
  pattern recognition}, pages 2161--2169, 2015.

\bibitem[Besl and McKay(1992)]{besl1992method}
Paul~J Besl and Neil~D McKay.
\newblock Method for registration of 3-d shapes.
\newblock In \emph{Sensor fusion IV: control paradigms and data structures},
  pages 586--606. Spie, 1992.

\bibitem[Bhattacharya and Govindu(2019)]{bhattacharya2019efficient}
Uttaran Bhattacharya and Venu~Madhav Govindu.
\newblock Efficient and robust registration on the 3d special euclidean group.
\newblock In \emph{Proceedings of the IEEE/CVF International Conference on
  Computer Vision}, pages 5885--5894, 2019.

\bibitem[Birdal and Ilic(2015)]{birdal2015point}
Tolga Birdal and Slobodan Ilic.
\newblock Point pair features based object detection and pose estimation
  revisited.
\newblock In \emph{2015 International conference on 3D vision}, pages 527--535.
  IEEE, 2015.

\bibitem[Birdal and Ilic(2017)]{birdal2017cad}
Tolga Birdal and Slobodan Ilic.
\newblock Cad priors for accurate and flexible instance reconstruction.
\newblock In \emph{Proceedings of the IEEE international conference on computer
  vision}, pages 133--142, 2017.

\bibitem[Birdal and Simsekli(2019)]{birdal2019probabilistic}
Tolga Birdal and Umut Simsekli.
\newblock Probabilistic permutation synchronization using the riemannian
  structure of the birkhoff polytope.
\newblock In \emph{Proceedings of the IEEE/CVF Conference on Computer Vision
  and Pattern Recognition}, pages 11105--11116, 2019.

\bibitem[Birdal et~al.(2018)Birdal, Simsekli, Eken, and
  Ilic]{birdal2018bayesian}
Tolga Birdal, Umut Simsekli, Mustafa~Onur Eken, and Slobodan Ilic.
\newblock Bayesian pose graph optimization via bingham distributions and
  tempered geodesic mcmc.
\newblock \emph{Advances in neural information processing systems}, 31, 2018.

\bibitem[Bustos and Chin(2017)]{bustos2017guaranteed}
Alvaro~Parra Bustos and Tat-Jun Chin.
\newblock Guaranteed outlier removal for point cloud registration with
  correspondences.
\newblock \emph{IEEE transactions on pattern analysis and machine
  intelligence}, 40\penalty0 (12):\penalty0 2868--2882, 2017.

\bibitem[Chatterjee and Govindu(2013)]{chatterjee2013efficient}
Avishek Chatterjee and Venu~Madhav Govindu.
\newblock Efficient and robust large-scale rotation averaging.
\newblock In \emph{Proceedings of the IEEE International Conference on Computer
  Vision}, pages 521--528, 2013.

\bibitem[Chen et~al.(2023)Chen, Liu, Li, Ding, and Feng]{chen2023deepmapping2}
Chao Chen, Xinhao Liu, Yiming Li, Li Ding, and Chen Feng.
\newblock Deepmapping2: Self-supervised large-scale lidar map optimization.
\newblock In \emph{Proceedings of the IEEE/CVF Conference on Computer Vision
  and Pattern Recognition}, pages 9306--9316, 2023.

\bibitem[Chetverikov et~al.(2002)Chetverikov, Svirko, Stepanov, and
  Krsek]{chetverikov2002trimmed}
Dmitry Chetverikov, Dmitry Svirko, Dmitry Stepanov, and Pavel Krsek.
\newblock The trimmed iterative closest point algorithm.
\newblock In \emph{2002 International Conference on Pattern Recognition}, pages
  545--548. IEEE, 2002.

\bibitem[Choi et~al.(2015)Choi, Zhou, and Koltun]{choi2015robust}
Sungjoon Choi, Qian-Yi Zhou, and Vladlen Koltun.
\newblock Robust reconstruction of indoor scenes.
\newblock In \emph{Proceedings of the IEEE conference on computer vision and
  pattern recognition}, pages 5556--5565, 2015.

\bibitem[Choy et~al.(2019)Choy, Park, and Koltun]{Choy2019FCGF}
Christopher Choy, Jaesik Park, and Vladlen Koltun.
\newblock Fully convolutional geometric features.
\newblock In \emph{ICCV}, 2019.

\bibitem[Cui and Tan(2015)]{cui2015global}
Zhaopeng Cui and Ping Tan.
\newblock Global structure-from-motion by similarity averaging.
\newblock In \emph{Proceedings of the IEEE International Conference on Computer
  Vision}, pages 864--872, 2015.

\bibitem[Deng et~al.(2018{\natexlab{a}})Deng, Birdal, and Ilic]{deng2018ppf}
Haowen Deng, Tolga Birdal, and Slobodan Ilic.
\newblock Ppf-foldnet: Unsupervised learning of rotation invariant 3d local
  descriptors.
\newblock In \emph{Proceedings of the European conference on computer vision},
  pages 602--618, 2018{\natexlab{a}}.

\bibitem[Deng et~al.(2018{\natexlab{b}})Deng, Birdal, and Ilic]{deng2018ppfnet}
Haowen Deng, Tolga Birdal, and Slobodan Ilic.
\newblock Ppfnet: Global context aware local features for robust 3d point
  matching.
\newblock In \emph{Proceedings of the IEEE conference on computer vision and
  pattern recognition}, pages 195--205, 2018{\natexlab{b}}.

\bibitem[Dhariwal and Nichol(2021)]{dhariwal2021diffusion}
Prafulla Dhariwal and Alexander Nichol.
\newblock Diffusion models beat gans on image synthesis.
\newblock \emph{Advances in neural information processing systems},
  34:\penalty0 8780--8794, 2021.

\bibitem[Drost et~al.(2010)Drost, Ulrich, Navab, and Ilic]{drost2010model}
Bertram Drost, Markus Ulrich, Nassir Navab, and Slobodan Ilic.
\newblock Model globally, match locally: Efficient and robust 3d object
  recognition.
\newblock In \emph{2010 IEEE computer society conference on computer vision and
  pattern recognition}, pages 998--1005. Ieee, 2010.

\bibitem[Endres et~al.(2013)Endres, Hess, Sturm, Cremers, and
  Burgard]{endres20133}
Felix Endres, J{\"u}rgen Hess, J{\"u}rgen Sturm, Daniel Cremers, and Wolfram
  Burgard.
\newblock 3-d mapping with an rgb-d camera.
\newblock \emph{IEEE transactions on robotics}, 30\penalty0 (1):\penalty0
  177--187, 2013.

\bibitem[Fantoni et~al.(2012)Fantoni, Castellani, and
  Fusiello]{fantoni2012accurate}
Simone Fantoni, Umberto Castellani, and Andrea Fusiello.
\newblock Accurate and automatic alignment of range surfaces.
\newblock In \emph{2012 Second International Conference on 3D Imaging,
  Modeling, Processing, Visualization \& Transmission}, pages 73--80. IEEE,
  2012.

\bibitem[Fischler and Bolles(1981)]{fischler1981random}
Martin~A Fischler and Robert~C Bolles.
\newblock Random sample consensus: a paradigm for model fitting with
  applications to image analysis and automated cartography.
\newblock \emph{Communications of the ACM}, 24\penalty0 (6):\penalty0 381--395,
  1981.

\bibitem[Flint et~al.(2007)Flint, Dick, and Van Den~Hengel]{flint2007thrift}
Alex Flint, Anthony Dick, and Anton Van Den~Hengel.
\newblock Thrift: Local 3d structure recognition.
\newblock In \emph{Biennial Conference of the Australian Pattern Recognition
  Society on Digital Image Computing Techniques and Applications}, pages
  182--188. IEEE, 2007.

\bibitem[Geiger et~al.(2012)Geiger, Lenz, and Urtasun]{geiger2012we}
Andreas Geiger, Philip Lenz, and Raquel Urtasun.
\newblock Are we ready for autonomous driving? the kitti vision benchmark
  suite.
\newblock In \emph{2012 IEEE conference on computer vision and pattern
  recognition}, pages 3354--3361. IEEE, 2012.

\bibitem[Gojcic et~al.(2019)Gojcic, Zhou, Wegner, and
  Wieser]{gojcic2019perfect}
Zan Gojcic, Caifa Zhou, Jan~D Wegner, and Andreas Wieser.
\newblock The perfect match: 3d point cloud matching with smoothed densities.
\newblock In \emph{Proceedings of the IEEE/CVF conference on computer vision
  and pattern recognition}, pages 5545--5554, 2019.

\bibitem[Gojcic et~al.(2020)Gojcic, Zhou, Wegner, Guibas, and
  Birdal]{gojcic2020learning}
Zan Gojcic, Caifa Zhou, Jan~D Wegner, Leonidas~J Guibas, and Tolga Birdal.
\newblock Learning multiview 3d point cloud registration.
\newblock In \emph{Proceedings of the IEEE/CVF conference on computer vision
  and pattern recognition}, pages 1759--1769, 2020.

\bibitem[Govindu(2004)]{govindu2004lie}
Venu~Madhav Govindu.
\newblock Lie-algebraic averaging for globally consistent motion estimation.
\newblock In \emph{Proceedings of the 2004 IEEE Computer Society Conference on
  Computer Vision and Pattern Recognition, 2004. CVPR 2004.}, pages I--I. IEEE,
  2004.

\bibitem[Hartley et~al.(2011)Hartley, Aftab, and Trumpf]{hartley2011l1}
Richard Hartley, Khurrum Aftab, and Jochen Trumpf.
\newblock L1 rotation averaging using the weiszfeld algorithm.
\newblock In \emph{CVPR 2011}, pages 3041--3048. IEEE, 2011.

\bibitem[Hinterstoisser et~al.(2016)Hinterstoisser, Lepetit, Rajkumar, and
  Konolige]{hinterstoisser2016going}
Stefan Hinterstoisser, Vincent Lepetit, Naresh Rajkumar, and Kurt Konolige.
\newblock Going further with point pair features.
\newblock In \emph{Computer Vision--ECCV 2016: 14th European Conference,
  Amsterdam, The Netherlands, October 11-14, 2016, Proceedings, Part III 14},
  pages 834--848. Springer, 2016.

\bibitem[Huang et~al.(2021)Huang, Gojcic, Usvyatsov, Wieser, and
  Schindler]{huang2021predator}
Shengyu Huang, Zan Gojcic, Mikhail Usvyatsov, Andreas Wieser, and Konrad
  Schindler.
\newblock Predator: Registration of 3d point clouds with low overlap.
\newblock In \emph{Proceedings of the IEEE/CVF Conference on computer vision
  and pattern recognition}, pages 4267--4276, 2021.

\bibitem[Huang et~al.(2017)Huang, Liang, Bajaj, and
  Huang]{huang2017translation}
Xiangru Huang, Zhenxiao Liang, Chandrajit Bajaj, and Qixing Huang.
\newblock Translation synchronization via truncated least squares.
\newblock \emph{Advances in neural information processing systems}, 30, 2017.

\bibitem[Huang et~al.(2019)Huang, Liang, Zhou, Xie, Guibas, and
  Huang]{huang2019learning}
Xiangru Huang, Zhenxiao Liang, Xiaowei Zhou, Yao Xie, Leonidas~J Guibas, and
  Qixing Huang.
\newblock Learning transformation synchronization.
\newblock In \emph{Proceedings of the IEEE/CVF conference on computer vision
  and pattern recognition}, pages 8082--8091, 2019.

\bibitem[Huber and Hebert(2003)]{huber2003fully}
Daniel~F Huber and Martial Hebert.
\newblock Fully automatic registration of multiple 3d data sets.
\newblock \emph{Image and Vision Computing}, 21\penalty0 (7):\penalty0
  637--650, 2003.

\bibitem[Jatavallabhula et~al.(2023)Jatavallabhula, Kuwajerwala, Gu, Omama,
  Chen, Li, Iyer, Saryazdi, Keetha, Tewari,
  et~al.]{jatavallabhula2023conceptfusion}
Krishna~Murthy Jatavallabhula, Alihusein Kuwajerwala, Qiao Gu, Mohd Omama, Tao
  Chen, Shuang Li, Ganesh Iyer, Soroush Saryazdi, Nikhil Keetha, Ayush Tewari,
  et~al.
\newblock Conceptfusion: Open-set multimodal 3d mapping.
\newblock 2023.

\bibitem[Jin et~al.(2023)Jin, Barath, Pollefeys, and Armeni]{jin2023q}
Shengze Jin, Daniel Barath, Marc Pollefeys, and Iro Armeni.
\newblock {Q-REG}: End-to-end trainable point cloud registration with surface
  curvature.
\newblock \emph{arXiv preprint arXiv:2309.16023}, 2023.

\bibitem[Johnson and Hebert(1999)]{johnson1999using}
Andrew~E Johnson and Martial Hebert.
\newblock Using spin images for efficient object recognition in cluttered 3d
  scenes.
\newblock \emph{IEEE Transactions on pattern analysis and machine
  intelligence}, 21\penalty0 (5):\penalty0 433--449, 1999.

\bibitem[Khoury et~al.(2017)Khoury, Zhou, and Koltun]{khoury2017learning}
Marc Khoury, Qian-Yi Zhou, and Vladlen Koltun.
\newblock Learning compact geometric features.
\newblock In \emph{Proceedings of the IEEE international conference on computer
  vision}, pages 153--161, 2017.

\bibitem[Li and Hartley(2007)]{li20073d}
Hongdong Li and Richard Hartley.
\newblock The 3d-3d registration problem revisited.
\newblock In \emph{2007 IEEE 11th international conference on computer vision},
  pages 1--8. IEEE, 2007.

\bibitem[Li and Ling(2021)]{li2021pogo}
Xinyi Li and Haibin Ling.
\newblock Pogo-net: pose graph optimization with graph neural networks.
\newblock In \emph{Proceedings of the IEEE/CVF International Conference on
  Computer Vision}, pages 5895--5905, 2021.

\bibitem[Lu et~al.(2019{\natexlab{a}})Lu, Wan, Zhou, Fu, Yuan, and
  Song]{lu2019deepvcp}
Weixin Lu, Guowei Wan, Yao Zhou, Xiangyu Fu, Pengfei Yuan, and Shiyu Song.
\newblock Deepvcp: An end-to-end deep neural network for point cloud
  registration.
\newblock In \emph{Proceedings of the IEEE/CVF international conference on
  computer vision}, pages 12--21, 2019{\natexlab{a}}.

\bibitem[Lu et~al.(2019{\natexlab{b}})Lu, Zhou, Wan, Hou, and Song]{lu2019l3}
Weixin Lu, Yao Zhou, Guowei Wan, Shenhua Hou, and Shiyu Song.
\newblock L3-net: Towards learning based lidar localization for autonomous
  driving.
\newblock In \emph{Proceedings of the IEEE/CVF Conference on Computer Vision
  and Pattern Recognition}, pages 6389--6398, 2019{\natexlab{b}}.

\bibitem[Maset et~al.(2017)Maset, Arrigoni, and Fusiello]{maset2017practical}
Eleonora Maset, Federica Arrigoni, and Andrea Fusiello.
\newblock Practical and efficient multi-view matching.
\newblock In \emph{Proceedings of the IEEE International Conference on Computer
  Vision}, pages 4568--4576, 2017.

\bibitem[Mellado et~al.(2014)Mellado, Aiger, and Mitra]{mellado2014super}
Nicolas Mellado, Dror Aiger, and Niloy~J Mitra.
\newblock Super 4pcs fast global pointcloud registration via smart indexing.
\newblock In \emph{Computer graphics forum}, pages 205--215. Wiley Online
  Library, 2014.

\bibitem[Mian et~al.(2006)Mian, Bennamoun, and Owens]{mian2006three}
Ajmal~S Mian, Mohammed Bennamoun, and Robyn Owens.
\newblock Three-dimensional model-based object recognition and segmentation in
  cluttered scenes.
\newblock \emph{IEEE transactions on pattern analysis and machine
  intelligence}, 28\penalty0 (10):\penalty0 1584--1601, 2006.

\bibitem[Palieri et~al.(2020)Palieri, Morrell, Thakur, Ebadi, Nash, Chatterjee,
  Kanellakis, Carlone, Guaragnella, and Agha-Mohammadi]{palieri2020locus}
Matteo Palieri, Benjamin Morrell, Abhishek Thakur, Kamak Ebadi, Jeremy Nash,
  Arghya Chatterjee, Christoforos Kanellakis, Luca Carlone, Cataldo
  Guaragnella, and Ali-akbar Agha-Mohammadi.
\newblock Locus: A multi-sensor lidar-centric solution for high-precision
  odometry and 3d mapping in real-time.
\newblock \emph{IEEE Robotics and Automation Letters}, 6\penalty0 (2):\penalty0
  421--428, 2020.

\bibitem[Purkait et~al.(2020)Purkait, Chin, and Reid]{purkait2020neurora}
Pulak Purkait, Tat-Jun Chin, and Ian Reid.
\newblock Neurora: Neural robust rotation averaging.
\newblock In \emph{European Conference on Computer Vision}, pages 137--154.
  Springer, 2020.

\bibitem[Qin et~al.(2022)Qin, Yu, Wang, Guo, Peng, and Xu]{qin2022geometric}
Zheng Qin, Hao Yu, Changjian Wang, Yulan Guo, Yuxing Peng, and Kai Xu.
\newblock Geometric transformer for fast and robust point cloud registration.
\newblock In \emph{CVPR}, 2022.

\bibitem[Qin et~al.(2023)Qin, Yu, Wang, Guo, Peng, Ilic, Hu, and
  Xu]{qin2023geotransformer}
Zheng Qin, Hao Yu, Changjian Wang, Yulan Guo, Yuxing Peng, Slobodan Ilic, Dewen
  Hu, and Kai Xu.
\newblock Geotransformer: Fast and robust point cloud registration with
  geometric transformer.
\newblock \emph{IEEE Transactions on Pattern Analysis and Machine
  Intelligence}, 2023.

\bibitem[Rabbani et~al.(2007)Rabbani, Dijkman, van~den Heuvel, and
  Vosselman]{rabbani2007integrated}
Tahir Rabbani, Sander Dijkman, Frank van~den Heuvel, and George Vosselman.
\newblock An integrated approach for modelling and global registration of point
  clouds.
\newblock \emph{ISPRS journal of Photogrammetry and Remote Sensing},
  61\penalty0 (6):\penalty0 355--370, 2007.

\bibitem[Rusu et~al.(2009{\natexlab{a}})Rusu, Blodow, and Beetz]{rusu2009FPFH}
Radu~Bogdan Rusu, Nico Blodow, and Michael Beetz.
\newblock Fast point feature histograms ({FPFH}) for 3{D} registration.
\newblock In \emph{ICRA}, 2009{\natexlab{a}}.

\bibitem[Rusu et~al.(2009{\natexlab{b}})Rusu, Blodow, and Beetz]{rusu2009fast}
Radu~Bogdan Rusu, Nico Blodow, and Michael Beetz.
\newblock Fast point feature histograms (fpfh) for 3d registration.
\newblock In \emph{IEEE international conference on robotics and automation},
  pages 3212--3217. IEEE, 2009{\natexlab{b}}.

\bibitem[Sarlin et~al.(2022)Sarlin, Dusmanu, Sch{\"o}nberger, Speciale, Gruber,
  Larsson, Miksik, and Pollefeys]{sarlin2022lamar}
Paul-Edouard Sarlin, Mihai Dusmanu, Johannes~L Sch{\"o}nberger, Pablo Speciale,
  Lukas Gruber, Viktor Larsson, Ondrej Miksik, and Marc Pollefeys.
\newblock Lamar: Benchmarking localization and mapping for augmented reality.
\newblock In \emph{European Conference on Computer Vision}, pages 686--704.
  Springer, 2022.

\bibitem[Steinbrucker et~al.(2013)Steinbrucker, Kerl, and
  Cremers]{steinbrucker2013large}
Frank Steinbrucker, Christian Kerl, and Daniel Cremers.
\newblock Large-scale multi-resolution surface reconstruction from rgb-d
  sequences.
\newblock In \emph{Proceedings of the IEEE International Conference on Computer
  Vision}, pages 3264--3271, 2013.

\bibitem[Sun et~al.(2023)Sun, Hao, Huang, Savarese, Schindler, Pollefeys, and
  Armeni]{nss2023}
Tao Sun, Yan Hao, Shengyu Huang, Silvio Savarese, Konrad Schindler, Marc
  Pollefeys, and Iro Armeni.
\newblock Nothing stands still: A spatiotemporal benchmark on 3d point cloud
  registration under large geometric and temporal change, 2023.

\bibitem[Taguchi et~al.(2013)Taguchi, Jian, Ramalingam, and
  Feng]{taguchi2013point}
Yuichi Taguchi, Yong-Dian Jian, Srikumar Ramalingam, and Chen Feng.
\newblock Point-plane slam for hand-held 3d sensors.
\newblock In \emph{2013 IEEE international conference on robotics and
  automation}, pages 5182--5189. IEEE, 2013.

\bibitem[Takmaz et~al.(2023)Takmaz, Fedele, Sumner, Pollefeys, Tombari, and
  Engelmann]{Takmaz2023openmask3d}
Ay{\c{c}}a Takmaz, Elisabetta Fedele, Robert~W Sumner, Marc Pollefeys, Federico
  Tombari, and Francis Engelmann.
\newblock {OpenMask3D: Open-Vocabulary 3D Instance Segmentation}.
\newblock In \emph{Conference on Neural Information Processing Systems
  (NeurIPS)}, 2023.

\bibitem[Tejus et~al.(2023)Tejus, Zara, Rota, Fusiello, Ricci, and
  Arrigoni]{tejus2023rotation}
Gk Tejus, Giacomo Zara, Paolo Rota, Andrea Fusiello, Elisa Ricci, and Federica
  Arrigoni.
\newblock Rotation synchronization via deep matrix factorization.
\newblock 2023.

\bibitem[Theiler et~al.(2014)Theiler, Wegner, and
  Schindler]{theiler2014keypoint}
Pascal~Willy Theiler, Jan~Dirk Wegner, and Konrad Schindler.
\newblock Keypoint-based 4-points congruent sets--automated marker-less
  registration of laser scans.
\newblock \emph{ISPRS journal of photogrammetry and remote sensing},
  96:\penalty0 149--163, 2014.

\bibitem[Theiler et~al.(2015)Theiler, Wegner, and
  Schindler]{theiler2015globally}
Pascal~Willy Theiler, Jan~Dirk Wegner, and Konrad Schindler.
\newblock Globally consistent registration of terrestrial laser scans via graph
  optimization.
\newblock \emph{ISPRS journal of photogrammetry and remote sensing},
  109:\penalty0 126--138, 2015.

\bibitem[Thomas et~al.(2019)Thomas, Qi, Deschaud, Marcotegui, Goulette, and
  Guibas]{thomas2019kpconv}
Hugues Thomas, Charles~R Qi, Jean-Emmanuel Deschaud, Beatriz Marcotegui,
  Fran{\c{c}}ois Goulette, and Leonidas~J Guibas.
\newblock Kpconv: Flexible and deformable convolution for point clouds.
\newblock In \emph{Proceedings of the IEEE/CVF international conference on
  computer vision}, pages 6411--6420, 2019.

\bibitem[Tombari et~al.(2010)Tombari, Salti, and Di~Stefano]{tombari2010unique}
Federico Tombari, Samuele Salti, and Luigi Di~Stefano.
\newblock Unique signatures of histograms for local surface description.
\newblock In \emph{European Conference on Computer Vision}, pages 356--369.
  Springer, 2010.

\bibitem[Torsello et~al.(2011)Torsello, Rodola, and
  Albarelli]{torsello2011multiview}
Andrea Torsello, Emanuele Rodola, and Andrea Albarelli.
\newblock Multiview registration via graph diffusion of dual quaternions.
\newblock In \emph{CVPR 2011}, pages 2441--2448. IEEE, 2011.

\bibitem[Wang et~al.(2023{\natexlab{a}})Wang, Liu, Dong, Guo, Liu, Wang, and
  Yang]{wang2023robust}
Haiping Wang, Yuan Liu, Zhen Dong, Yulan Guo, Yu-Shen Liu, Wenping Wang, and
  Bisheng Yang.
\newblock Robust multiview point cloud registration with reliable pose graph
  initialization and history reweighting.
\newblock In \emph{Proceedings of the IEEE/CVF Conference on Computer Vision
  and Pattern Recognition}, pages 9506--9515, 2023{\natexlab{a}}.

\bibitem[Wang et~al.(2023{\natexlab{b}})Wang, Rupprecht, and
  Novotny]{wang2023posediffusion}
Jianyuan Wang, Christian Rupprecht, and David Novotny.
\newblock Posediffusion: Solving pose estimation via diffusion-aided bundle
  adjustment.
\newblock In \emph{Proceedings of the IEEE/CVF International Conference on
  Computer Vision}, pages 9773--9783, 2023{\natexlab{b}}.

\bibitem[Wang and Solomon(2019)]{wang2019deep}
Yue Wang and Justin~M Solomon.
\newblock Deep closest point: Learning representations for point cloud
  registration.
\newblock In \emph{Proceedings of the IEEE/CVF international conference on
  computer vision}, pages 3523--3532, 2019.

\bibitem[Whelan et~al.(2013)Whelan, Kaess, Leonard, and
  McDonald]{whelan2013deformation}
Thomas Whelan, Michael Kaess, John~J Leonard, and John McDonald.
\newblock Deformation-based loop closure for large scale dense rgb-d slam.
\newblock In \emph{2013 IEEE/RSJ International Conference on Intelligent Robots
  and Systems}, pages 548--555. IEEE, 2013.

\bibitem[Wolcott and Eustice(2014)]{wolcott2014visual}
Ryan~W Wolcott and Ryan~M Eustice.
\newblock Visual localization within lidar maps for automated urban driving.
\newblock In \emph{2014 IEEE/RSJ International Conference on Intelligent Robots
  and Systems}, pages 176--183. IEEE, 2014.

\bibitem[Xiao et~al.(2013)Xiao, Owens, and Torralba]{xiao2013sun3d}
Jianxiong Xiao, Andrew Owens, and Antonio Torralba.
\newblock Sun3d: A database of big spaces reconstructed using sfm and object
  labels.
\newblock In \emph{Proceedings of the IEEE international conference on computer
  vision}, pages 1625--1632, 2013.

\bibitem[Yang et~al.(2015)Yang, Li, Campbell, and Jia]{yang2015go}
Jiaolong Yang, Hongdong Li, Dylan Campbell, and Yunde Jia.
\newblock Go-icp: A globally optimal solution to 3d icp point-set registration.
\newblock \emph{IEEE transactions on pattern analysis and machine
  intelligence}, 38\penalty0 (11):\penalty0 2241--2254, 2015.

\bibitem[Yang et~al.(2021)Yang, Li, Rahim, Cui, and Tan]{yang2021end}
Luwei Yang, Heng Li, Jamal~Ahmed Rahim, Zhaopeng Cui, and Ping Tan.
\newblock End-to-end rotation averaging with multi-source propagation.
\newblock In \emph{Proceedings of the IEEE/CVF Conference on Computer Vision
  and Pattern Recognition}, pages 11774--11783, 2021.

\bibitem[Yew and Lee(2018)]{yew20183dfeat}
Zi~Jian Yew and Gim~Hee Lee.
\newblock 3dfeat-net: Weakly supervised local 3d features for point cloud
  registration.
\newblock In \emph{Proceedings of the European conference on computer vision},
  pages 607--623, 2018.

\bibitem[Yew and Lee(2021)]{yew2021learning}
Zi~Jian Yew and Gim~Hee Lee.
\newblock Learning iterative robust transformation synchronization.
\newblock In \emph{2021 International Conference on 3D Vision (3DV)}, pages
  1206--1215. IEEE, 2021.

\bibitem[Yew and Lee(2022)]{yew2022regtr}
Zi~Jian Yew and Gim~Hee Lee.
\newblock Regtr: End-to-end point cloud correspondences with transformers.
\newblock In \emph{CVPR}, 2022.

\bibitem[Yu et~al.(2023{\natexlab{a}})Yu, Qin, Hou, Saleh, Li, Busam, and
  Ilic]{yu2023rotation}
Hao Yu, Zheng Qin, Ji Hou, Mahdi Saleh, Dongsheng Li, Benjamin Busam, and
  Slobodan Ilic.
\newblock Rotation-invariant transformer for point cloud matching.
\newblock In \emph{Proceedings of the IEEE/CVF Conference on Computer Vision
  and Pattern Recognition}, pages 5384--5393, 2023{\natexlab{a}}.

\bibitem[Yu et~al.(2023{\natexlab{b}})Yu, Ren, Zhang, Zhou, Lin, and
  Dai]{yu2023peal}
Junle Yu, Luwei Ren, Yu Zhang, Wenhui Zhou, Lili Lin, and Guojun Dai.
\newblock Peal: Prior-embedded explicit attention learning for low-overlap
  point cloud registration.
\newblock In \emph{Proceedings of the IEEE/CVF Conference on Computer Vision
  and Pattern Recognition}, pages 17702--17711, 2023{\natexlab{b}}.

\bibitem[Zeisl et~al.(2013)Zeisl, Koser, and Pollefeys]{zeisl2013automatic}
Bernhard Zeisl, Kevin Koser, and Marc Pollefeys.
\newblock Automatic registration of rgb-d scans via salient directions.
\newblock In \emph{Proceedings of the IEEE international conference on computer
  vision}, pages 2808--2815, 2013.

\bibitem[Zeng et~al.(2017)Zeng, Song, Nie{\ss}ner, Fisher, Xiao, and
  Funkhouser]{zeng20173dmatch}
Andy Zeng, Shuran Song, Matthias Nie{\ss}ner, Matthew Fisher, Jianxiong Xiao,
  and Thomas Funkhouser.
\newblock 3dmatch: Learning local geometric descriptors from rgb-d
  reconstructions.
\newblock In \emph{Proceedings of the IEEE conference on computer vision and
  pattern recognition}, pages 1802--1811, 2017.

\bibitem[Zhou and Koltun(2013)]{zhou2013dense}
Qian-Yi Zhou and Vladlen Koltun.
\newblock Dense scene reconstruction with points of interest.
\newblock \emph{ACM Transactions on Graphics (ToG)}, 32\penalty0 (4):\penalty0
  1--8, 2013.

\bibitem[Zhou et~al.(2016)Zhou, Park, and Koltun]{zhou2016fast}
Qian-Yi Zhou, Jaesik Park, and Vladlen Koltun.
\newblock Fast global registration.
\newblock In \emph{Computer Vision--ECCV 2016: 14th European Conference,
  Amsterdam, The Netherlands, October 11-14, 2016, Proceedings, Part II 14},
  pages 766--782. Springer, 2016.

\end{thebibliography}
}
\clearpage
\setcounter{page}{1}
\maketitlesupplementary

\begin{abstract}
\noindent
In the supplemental material, we provide additional details about the following:\\
(I) Visualizations for pairwise registration on the 3DMatch and 3DLoMatch datasets (Section~\ref{sec:pairwise_visual}),\\
(II) Visualizations for multiway registration on the NSS dataset (Section~\ref{sec:multiway_visual}),\\
(III) Registration recall for multiway registration on the four datasets (Section~\ref{sec:registration_recall}),\\
(IV) Ablation study on pairwise registration results on the NSS dataset (Section~\ref{sec:more_nss}),\\
(V) The run-times of the methods (Section~\ref{sec:run_time}), and \\ 
(VI) A \href{https://youtu.be/dnzhKfPIoWg} {video} that gives a summary of our method and results.
\end{abstract}

\begin{table*}[h]
    \footnotesize
    \begin{center}
    \resizebox{1.0\linewidth}{!}{
    \begin{tabular}{l|ccc|l|ccc|ccc|ccc}
        \toprule
        \multirow{2}{*}{Method} & \multicolumn{3}{c|}{NSS} & & \multicolumn{3}{c|}{3DMatch} & \multicolumn{3}{c|}{3DLoMatch} & \multicolumn{3}{c}{KITTI} \\ 
        & RR (\%)$\uparrow$ & RE ($^{\circ}$)$\downarrow$ & TE (m)$\downarrow$ & & RR (\%)$\uparrow$ & RE ($^{\circ}$)$\downarrow$ & TE (cm)$\downarrow$ & RR (\%)$\uparrow$ & RE ($^{\circ}$)$\downarrow$ & TE (cm)$\downarrow$& RR (\%)$\uparrow$ & RE ($^{\circ}$)$\downarrow$ & TE (cm)$\downarrow$ \\
        \midrule
        Predator & 64.6 & 13.43 & 0.65 & PEAL & 94.1 & 4.72 & 15.8 & 78.8 & 16.03 & 50.2 & 75.7 & 9.46 & 11.85 \\
        {+ } Open3d~\cite{choi2015robust} & 51.3 & 12.76 & 0.64 & + Open3d & 81.8 & 4.72 & 15.8 & 68.9 & 14.23 & 45.1 & 83.2 & 6.21 & \phantom{1}7.72 \\
        {+ } DeepMapping2~\cite{chen2023deepmapping2} & 60.1 & 11.54 & 0.64 & + DeepM.\ & 82.7 & 4.23 & 14.5 & 70.1 & 13.25 & 39.4 & 91.5 & 3.34 & \phantom{1}6.04 \\
        {+ } LMPR~\cite{gojcic2020learning} & 65.1 & 11.35 & 0.62 & + LMPR & 82.4 & 3.98 & 12.6 & 70.7 & 13.07 & 37.3 & 81.1 & 6.79 & \phantom{1}7.89 \\
        {+ } LIRTS~\cite{yew2021learning} & 65.9 & 11.42 & 0.61 & + LIRTS & 86.9 & 3.95 & 12.0 & 76.2 & 11.52 & 36.0 & 84.7 & 5.17 & \phantom{1}6.94  \\
        {+ } RMPR~\cite{wang2023robust} & 66.9 & 10.87 & 0.62 & + RMPR & 95.9 & 3.57 & 11.6 & 83.1 & 10.18 & 34.4 & 86.1 & 4.69 & \phantom{1}6.38 \\
        {+ } \textbf{Wednesday} & 75.6 & \phantom{1}2.24 & 0.51 & + \textbf{Wednesday} & 96.8 & 2.58 & \phantom{1}9.4 & 86.4 & \phantom{1}7.21 & 29.1 & 94.6 & 2.52 & \phantom{1}5.92 \\
        \textbf{ODIN} + \textbf{Wednesday} & \textbf{78.3} & \phantom{1}\textbf{2.01} & \textbf{0.42} & & \textbf{97.3} & \textbf{2.32} & \phantom{1}\textbf{8.4} & \textbf{87.1} & \phantom{1}\textbf{6.44} & \textbf{26.5} & \textbf{96.2} & \textbf{2.18} & \phantom{1}\textbf{4.76} \\
        \bottomrule
    \end{tabular}
    }
    \end{center}
    \vspace{-5mm}
\caption{\textbf{Multiway point cloud registration} on the NSS~\cite{nss2023}, 3DMatch~\cite{zeng20173dmatch}, 3DLoMatch~\cite{huang2021predator} and KITTI~\cite{geiger2012we} datasets. 
The reported metrics are the registration recall (RR), average rotation (RE) and translation errors (TE). 
For each dataset, we choose the best-performing pairwise estimator from the baselines. 
We run Predator~\cite{huang2021predator} on NSS and PEAL~\cite{yu2023peal} on the other datasets. The best results are in \textbf{bold}. }
\label{tab:mr_rr}
\end{table*}

\begin{table*}[h]
    \footnotesize
    \begin{center}
    \resizebox{\linewidth}{!}{
    \begin{tabular}{l|ccc|ccc|ccc}
        \toprule
        \multirow{2}{*}{Method} & \multicolumn{3}{c|}{All spatiotemporal pairs} & \multicolumn{3}{c|}{Only same-stage pairs} & \multicolumn{3}{c}{Only different-stage pairs} \\ 
        & RR (\%)$\uparrow$ & RTE (m)$\downarrow$ & RRE (°)$\downarrow$ & RR (\%)$\uparrow$ & RTE (m)$\downarrow$ & RRE (°)$\downarrow$ & RR (\%)$\uparrow$ & RTE (m)$\downarrow$ & RRE (°)$\downarrow$ \\
        \midrule
        FPFH~\cite{rusu2009FPFH} & 11.70 & 2.23 & 45.32 & 30.82 & 2.42 & 29.35 & \phantom{1}0.42 & 4.06 & 78.01\\
        FCGF~\cite{Choy2019FCGF} & 24.43 & 2.04 & 39.89 & 42.86 & 2.23 & 32.12 & 10.52 & 3.23 & 53.24\\
        D3Feat~\cite{bai2020d3feat} & 22.73 & 2.26 & 33.09 & 36.51 & 2.05 & 27.22 & \phantom{1}4.76 & 2.53 & 40.76\\
        Predator~\cite{huang2021predator} & 64.97 & 0.65 & 13.52 & 92.99 & 0.27 & \phantom{1}4.83 & 28.42 & 1.16 & 24.85\\        GeoTransformer~\cite{qin2023geotransformer} & 39.07 & 0.99 & 22.93 & 55.59 & 0.73 & 17.02 & 17.51 & 1.34 & 30.62\\
        PEAL~\cite{yu2023peal} & 58.72 & 0.71 & 15.78 & 88.63 & 0.32 & \phantom{1}5.32 & 19.71 & 1.22 & 29.42 \\
        \textbf{ODIN} & \textbf{69.73} & \textbf{0.54} & \textbf{11.96} & \textbf{95.46} & \textbf{0.21} & \phantom{1}\textbf{4.36} & \textbf{36.17} & \textbf{0.97} & \textbf{21.87} \\
        \bottomrule
    \end{tabular}
    }
    \end{center}
    \vspace{-5mm}
    \caption{\textbf{Pairwise point cloud registration} on the NSS dataset. The reported metrics are the Registration Recall (RR), which measures the fraction of successfully registered pairs; the Relative Rotation Error (RRE); and the Relative Translation Error (RTE). We show ablation results for same-stage and different-stage pairs. The best results are in \textbf{bold}.}
    \label{tab:nss_more}
\end{table*}

\section{Visualizations of Pairwise Registration}
\label{sec:pairwise_visual}

In Figures~\ref{fig:qual_3D} and \ref{fig:qual_3DLo}, we show pairwise registration results on the \textit{3DLoMatch} and \textit{3DMatch} datasets, respectively. 
We do not show pairwise results on NSS and KITTI since: (i) the NSS point clouds represent spaces with ceiling information and lack details -- such pairwise registration results are hard to interpret even with the ceilings are cut off; and (ii)
the results on KITTI are quite saturated with all methods achieving good results.
While our ODIN achieves the most accurate registrations (as per Table 1 in the main paper), no significant difference is visible in the pairwise visualizations. 
We show the results of the three best matchers (according to Table 1 in the main paper), namely ODIN, PEAL~\cite{yu2023peal} and GeoTransformer~\cite{qin2022geometric}.

\subsection{Visualizations on 3DMatch}
\label{sec:3dmatch}

\begin{figure*}[p]
    \centering
    \includegraphics[width=0.99\linewidth]{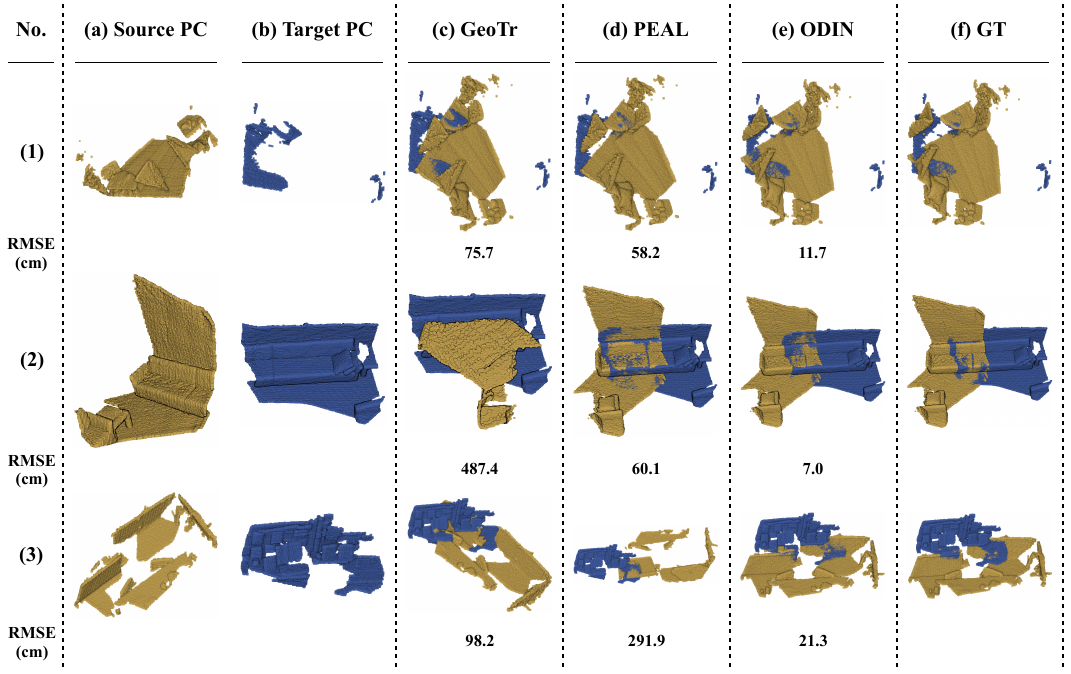}
    \caption{\textbf{Qualitative Results for the \textit{3DMatch~\cite{zeng20173dmatch}} dataset.} See Section~\ref{sec:3dmatch} for an explanation of the results. \textit{Best viewed in screen.}}
    \label{fig:qual_3D}
\end{figure*}

In Figure~\ref{fig:qual_3D}, we show examples of point cloud registration on the \textit{3DMatch} dataset. 
We also report the RMSE for all results, which we use to determine if two point clouds are correctly registered. 
Specifically, per row:

\vspace{1mm}\noindent\textbf{Row (1):} 
This example showcases a particularly challenging pair with a very low overlap in the point clouds. 
While all methods manage to estimate the correct pose coarsely, both GeoTransformer and PEAL achieve high RMSE.
The output of our ODIN is close to the ground truth transformation, with an RMSE that is substantially lower than its competitors.

\vspace{1mm}\noindent\textbf{Row (2):} 
In this case, GeoTransformer fails to find a correct pose, even coarsely. 
Similar to row (1), PEAL manages to output an acceptable transformation. However, it has a high RMSE, that is far from the ground truth.
The registration output of ODIN is almost an order of magnitude more accurate than that of PEAL in terms of RMSE, and, visually, it is very close to the ground truth registration. 
This success underscores the efficacy of our dual-stream architecture combined with the attention mechanism, which directs the network's focus towards regions of high confidence for more dependable correspondence inference. 
Additionally, the diffusion model plays a crucial role in eliminating noisy matches, further enhancing the overall precision.

\vspace{1mm}\noindent\textbf{Row (3):} In this example, both GeoTransformer and PEAL fail. ODIN has a higher RMSE than in the above examples, however, the registration output is visually acceptable and closer to the ground truth. 

While the examples show that there is still room for improvement, ODIN clearly achieves substantially better registrations than the state of the art in 3D Match. 

\subsection{Visualizations on 3DLoMatch}
\label{sec:3dlomatch}

\begin{figure*}[p]
    \centering
    \includegraphics[width=0.99\linewidth]{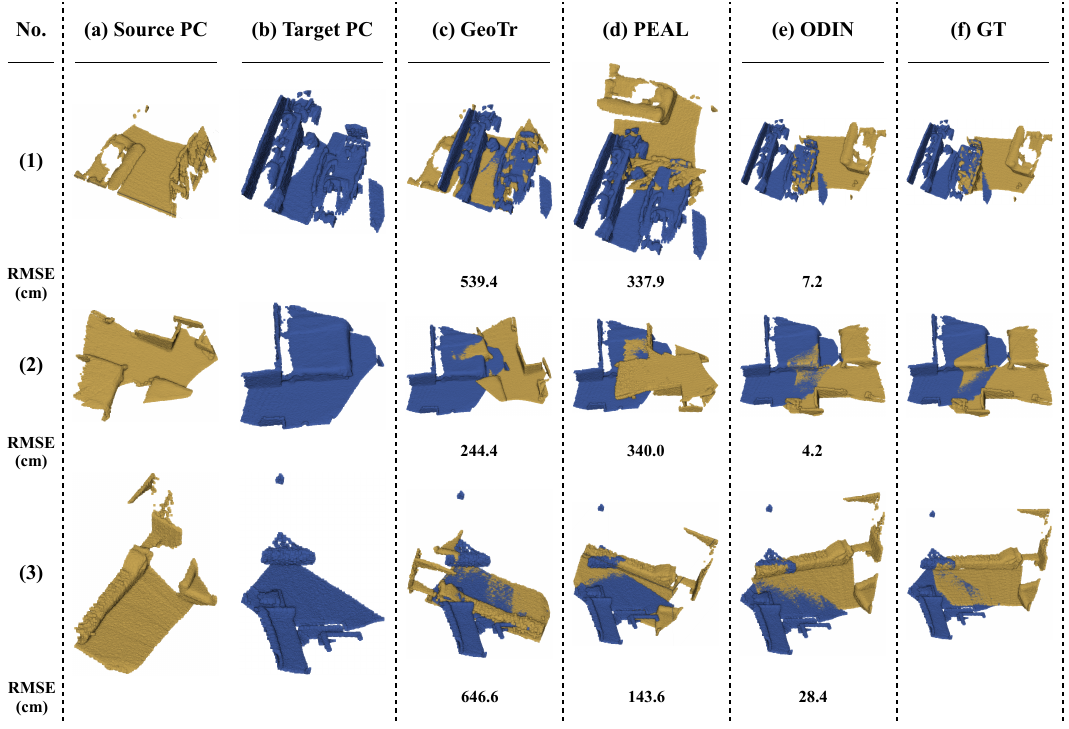}
    \caption{\textbf{Qualitative Results for the \textit{3DLoMatch~\cite{huang2021predator}} dataset.} See Section~\ref{sec:3dlomatch} for an explanation of the results. \textit{Best viewed in screen.}}
    \label{fig:qual_3DLo}
\end{figure*}

In Figure~\ref{fig:qual_3DLo}, we show examples of point cloud registration on the \textit{3DLoMatch} dataset. Similar to 3DMatch, we report the RMSE for all results. Specifically, per row:

\vspace{1mm}\noindent\textbf{Row (1):} In this example, we observe that ODIN recovers the pose very accurately, while both GeoTransformer and PEAL fail entirely. Their RMSE is two orders of magnitude higher than that of ODIN.  
This again highlights the importance of the proposed two-stream architecture and the diffusion-based denoising. 

\vspace{1mm}\noindent\textbf{Row (2):} Here, ODIN provides a close-to-GT pose. GeoTransformer and PEAL struggle to find a good pose.

\vspace{1mm}\noindent\textbf{Row (3):} In this example, all methods fail to recover a good pose. However, ODIN still manages a significantly lower RMSE than the other methods.
In addition, visually, the output is not far from the ground truth alignment.

\section{Visualizations of Multiway Registration}
\label{sec:multiway_visual}

\begin{figure*}[p]
    \centering
    \includegraphics[width=0.99\linewidth]{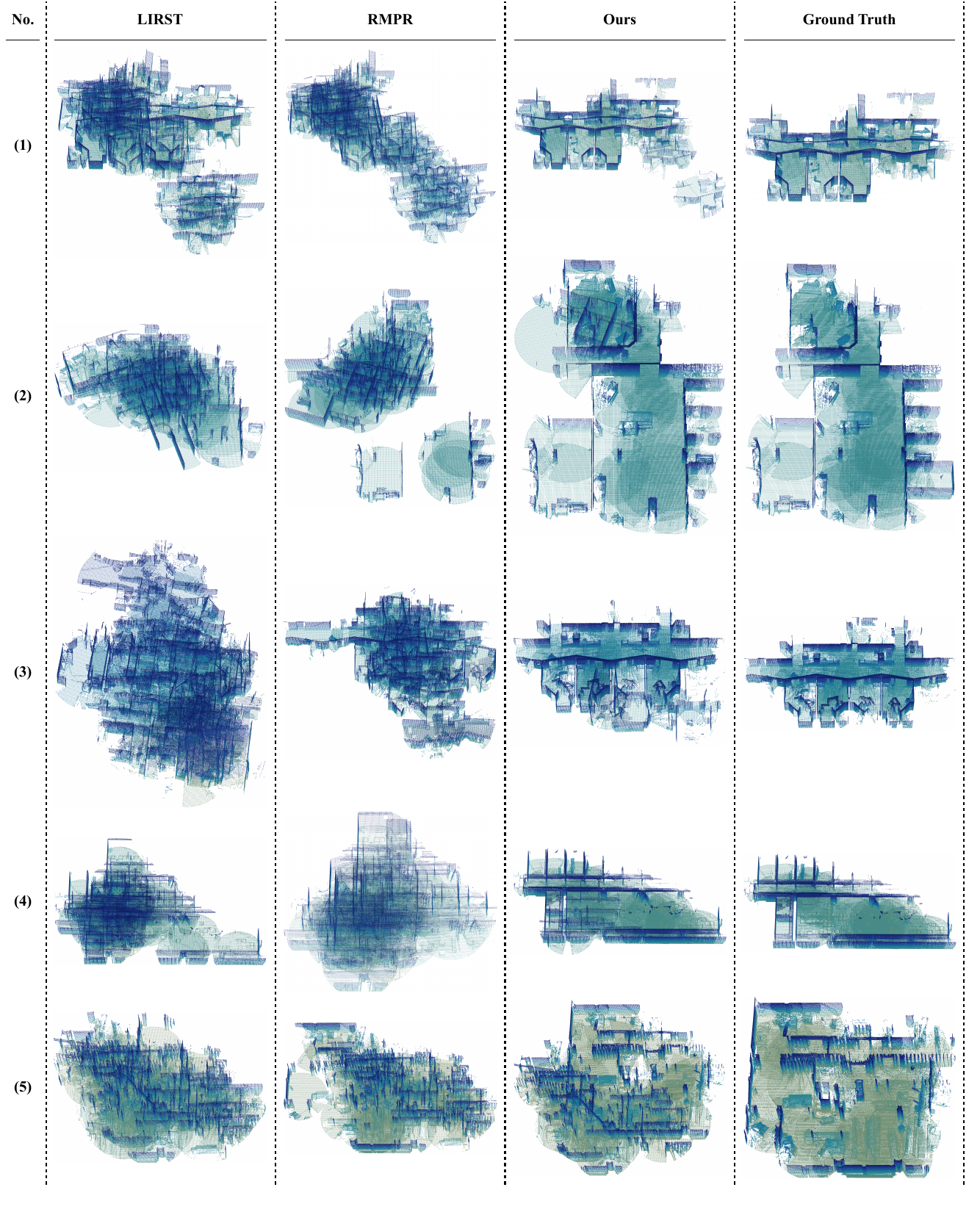}
    \caption{\textbf{Qualitative Results for the \textit{NSS~\cite{nss2023}} dataset.} See Section~\ref{sec:multiway_visual} for an explanation of the results. \textit{Best viewed in screen.}}
    \label{fig:qual_NSS}
\end{figure*}

In Figure~\ref{fig:qual_NSS}, we show examples of point cloud multiway registration for the \textit{NSS} dataset. We choose to visualize NSS as it is the most challenging dataset. 
We show results of our proposed method and \cite{yew2021learning,wang2023robust}. 
We choose \cite{yew2021learning,wang2023robust} as they are -- after ours -- the next best-performing methods as per Table 2 in the main paper. Specifically, per row:

\vspace{1mm}\noindent\textbf{Rows (1), (2), (3) and (4):} 
LIRST~\cite{yew2021learning} and RMPR~\cite{wang2023robust} fail to achieve an acceptable global registration in these examples. Their outputs are incomprehensible and are far from the expected results. Such outputs are frequent for these methods on this dataset. 
While our proposed method has inaccuracies, it provides substantially more accurate registrations that are not far from the ground truth. 
This highlights that the proposed multiway registration pipeline is more robust to such complicated scenarios than the state of the art. 

\vspace{1mm}\noindent\textbf{Row (5):} In this example, all methods fail to achieve a good registration. 
As before, both LIRST and RMPR results are incomprehensible. 
Our method manages to find the structure coarsely, however, there are mistakes, showing that there is still room for improvement.

\section{Multiway Registration Recall}
\label{sec:registration_recall}

We provide the registration recall (RR) for multiway registration on the NSS~\cite{nss2023}, 3DMatch~\cite{zeng20173dmatch}, 3DLoMatch~\cite{huang2021predator}, and KITTI~\cite{geiger2012we} datasets in Table~\ref{tab:mr_rr}.
RE and TE are taken from Table \textcolor{red}{2} in the main paper.
The successfully registered pairs are defined following the protocol from~\cite{nss2023, wang2023robust, qin2023geotransformer}. 
For each dataset, we choose the best-performing pairwise estimator from the baselines. We run Predator~\cite{huang2021predator} on NSS and PEAL~\cite{yu2023peal} on the other datasets. 

Our proposed \textbf{Wednesday} (without ODIN) consistently improves upon \textit{all} state-of-the-art algorithms and gains 0.9\% to 8.7\% in RR compared to the second-best method on the four datasets. Our full pipeline, \textbf{ODIN} + \textbf{Wednesday}, achieves additional improvements on RR on all datasets.

\section{Pairwise Ablation on the NSS Dataset}
\label{sec:more_nss}

We show an ablation of the pairwise point cloud registration results on the NSS dataset in Table~\ref{tab:nss_more}. The reported metrics are the Registration Recall (RR), which measures the fraction of successfully registered pairs; the Relative Rotation Error (RRE) in degrees ($^\circ$); and the Relative Translation Error (RTE) in meters ($m$). 
Specifically, we ablate the results for same-stage pairs and different-stage pairs as defined in the original paper~\cite{nss2023}, which evaluates independently the performance of pairs of point clouds from the same (w/o change) or different (w/ change) temporal stages. 

The first three columns (first block) show the results on all pairs regardless of being from same or different stages. 
This is the same as in Table 1 in the main paper.
For the same-stage pairs (second block), the results improve for all methods compared to the previous case and follow the same trend in the order of performance.
However, only three algorithms provide very high performance (above 88\%; Predator, PEAL, and ODIN), with ODIN being the most accurate. The rest follow at below 55\% of performance. 
On the different-stage pairs, there is a substantial difference (\ie, 7.7\% RR) between ODIN and the second best method, Predator. 
While ODIN is significantly better than all competitors, it is important to note that its RR on the different-stage pairs is still far from 100$\%$.
This highlights that further improvements are needed to robustly solve such complicated scenarios exhibiting temporal changes. 

We find it interesting that, while PEAL falls short compared to Predator, our ODIN significantly outperforms both, while building on similar architectural blocks as PEAL. 
This demonstrates the importance of the proposed two-stream attention learning architecture coupled with the diffusion denoising module. 
PEAL's effectiveness heavily relies on the initial pose provided by the GeoTransformer. 
It struggles to correct this initial pose if it is too inaccurate. In contrast, our method does not rely on an initial pose and, thus, it identifies correct correspondences more robustly.
It effectively filters out erroneous correspondences, retaining only those with high confidence.

\section{Processing Times}
\label{sec:run_time}

We evaluate the runtime on a computer with Intel(R) Xeon(R) CPU E3-1284L v4 @ 2.90GHz and GeForce RTX 3090 GPU. 
In Table~\ref{tab:PR_time}, we provide the total and average times in seconds of pairwise registration methods on the 3DMatch dataset. 
The total time represents the cumulative runtime for pairwise registration across the entire scene, while the average time denotes the mean duration expended for each individual pair.
The results show that the proposed ODIN runs at a similar speed to its less accurate alternatives. Specifically, it is marginally slower than GeoTransformer and PEAL, and it is twice as fast as Predator. 

In Table~\ref{tab:MR_time}, we provide the total and average times of multiway registration methods on the 3DMatch dataset. 
For this experiment, we compare using the same methods as those listed in Table \textcolor{red}{2} of the main paper.
The proposed method, Wednesday, falls in the middle in terms of runtime. 

In conclusion, there is no trade-off when using the proposed ODIN and Wednesday. They obtain state-of-the-art results while running at a similar speed as the baselines. 

\begin{table}[t]
    \footnotesize
    \begin{center}
    \begin{tabular}{l|cc}
        \toprule
        Method & Total Time (s) & Average Time (s)\\
        \midrule
        Predator & 460 & 0.26 \\
        GeoTr. & 159 & 0.09 \\
        PEAL & 212 & 0.12 \\
        ODIN & 248 & 0.14 \\
        \bottomrule
    \end{tabular}%
    \caption{Total and average time of pairwise point cloud registration pipelines on the 3DMatch dataset.}
    \label{tab:PR_time}
    \end{center}
\end{table}

\begin{table}[t]
    \footnotesize
    \begin{center}
    \begin{tabular}{l|cc}
        \toprule
        Method & Total Time (s) & Average Time (s)\\
        \midrule
        PEAL & 212 & 0.12 \\
        PEAL + Open3d & 283 & 0.16 \\
        PEAL + DeepMapping2 & 7399 & 4.18 \\
        PEAL + LMPR & 301 & 0.17 \\
        PEAL + LIRTS & 425 & 0.24 \\
        PEAL + RMPR & 244 & 0.14 \\
        PEAL + Wednesday  & 389 & 0.22 \\
        ODIN + Wednesday  & 425 & 0.24 \\
        \bottomrule
    \end{tabular}
\caption{Total time and average time per point cloud pair of multiway point cloud registration pipelines on the 3DMatch dataset.}
\label{tab:MR_time}
    \end{center}
\end{table}

\end{document}